\documentclass[runningheads]{llncs}

\usepackage{newpxtext,newpxmath}

\usepackage{graphicx}
\usepackage{color}
\usepackage[table,dvipsnames]{xcolor}
\usepackage[normalem]{ulem}
\usepackage{array}
\usepackage{changepage}
\usepackage{multicol}
\usepackage{ragged2e}
\usepackage{pdflscape}
\usepackage{tabularray}
\usepackage{booktabs}
\usepackage[longtable]{multirow}
\usepackage{longtable}
\usepackage{hyperref}
\usepackage{wrapfig}
\usepackage{xspace}

\usepackage{caption}
\usepackage{subcaption}
\usepackage{geometry}
\definecolor{mydarkblue}{rgb}{0,0.08,0.45}
\hypersetup{ %
    pdftitle={},
    pdfkeywords={},
    pdfborder=0 0 0,
    pdfpagemode=UseNone,
    colorlinks=true,
    linkcolor=mydarkblue,
    citecolor=mydarkblue,
    filecolor=mydarkblue,
    urlcolor=mydarkblue,
}

\usepackage{natbib}
\usepackage[parfill]{parskip}

\makeatletter
\renewcommand\subsubsection{\@startsection{subsubsection}{3}{\z@}%
                       {-18\p@ \@plus -4\p@ \@minus -4\p@}%
                       {0.5em \@plus 0.22em \@minus 0.1em}%
                       {\normalfont\normalsize\bfseries\boldmath}}

\renewcommand\paragraph{\@startsection{paragraph}{4}{\z@}%
            {-2.5ex\@plus -1ex \@minus -.25ex}%
            {1.25ex \@plus .25ex}%
            {\normalfont\normalsize\bfseries}}
\makeatother
\newcommand{\positiveImpact}[1]{+ #1}
\newcommand{\negativeImpact}[1]{- #1}

\definecolor{LFive}{rgb}{0.278, 0.365, 0.125}
\definecolor{LFour}{rgb}{0.584, 0.494, 0.039}
\definecolor{LThree}{rgb}{0.773, 0.361, 0.063}
\definecolor{LTwo}{rgb}{0.757, 0.51, 0.047}
\definecolor{LOne}{rgb}{0.753, 0.0, 0.0}
\definecolor{LNA}{rgb}{0.263, 0.263, 0.263}
\definecolor{LUnknown}{rgb}{0.6, 0.6, 0.6}
\definecolor{CompanyColor}{rgb}{0.851, 0.824, 0.914}
\definecolor{NonProfitColor}{rgb}{0.918, 0.82, 0.863}
\definecolor{GovernmentColor}{rgb}{0.812, 0.886, 0.953}

\newcommand{\CFiveTable}{\color{white}\cellcolor{LFive}C5}
\newcommand{\CFourTable}{\color{white}\cellcolor{LFour}C4}
\newcommand{\CThreeTable}{\color{white}\cellcolor{LThree}C3}
\newcommand{\CTwoTable}{\color{white}\cellcolor{LTwo}C2}
\newcommand{\COneTable}{\color{white}\cellcolor{LOne}C1}

\newcommand{\DFiveTable}{\color{white}\cellcolor{LFive}D5}
\newcommand{\DFourTable}{\color{white}\cellcolor{LFour}D4}
\newcommand{\DThreeTable}{\color{white}\cellcolor{LThree}D3}
\newcommand{\DTwoTable}{\color{white}\cellcolor{LTwo}D2}
\newcommand{\DOneTable}{\color{white}\cellcolor{LOne}D1}

\newcommand{\NATable}{\color{white}\cellcolor{LNA}N/A}
\newcommand{\UnkTable}{\color{white}\cellcolor{LUnknown}?}

\newcommand{\CFiveText}{{\color{LFive}C5}\xspace}

\newcommand{\COneText}{{\color{LOne}C1}\xspace}

\newcommand{\DFiveText}{{\color{LFive}D5}\xspace}

\newcommand{\DOneText}{{\color{LOne}D1}\xspace}

\newcommand{\midrulenospacing}{\specialrule{.4pt}{2pt}{0pt}}
\newcommand{\bottomrulenospacing}{\specialrule{.8pt}{0pt}{2pt}}

\newcommand{\DevCompany}[1]{\cellcolor{CompanyColor}#1}
\newcommand{\DevNonProfit}[1]{\cellcolor{NonProfitColor}#1}
\newcommand{\DevGovernment}[1]{\cellcolor{GovernmentColor}#1}

\definecolor{RID}{rgb}{0.329, 0.51, 0.208}
\definecolor{SS}{rgb}{0.773, 0.353, 0.067}
\definecolor{EAU}{rgb}{0.749, 0.565, 0.008}
\definecolor{BSA}{rgb}{0.176, 0.459, 0.714}

\newcommand{\RIDText}[1]{\color{RID}#1}

\newcommand{\SSText}[1]{\color{SS}#1}
\newcommand{\EAUText}[1]{\color{EAU}#1}
\newcommand{\BSAText}[1]{\color{BSA}#1}

\begin{document}
\title{Risks and Opportunities of Open-Source Generative AI}

\author{
    Francisco Eiras\inst{1} \and
    Aleksandar Petrov\inst{1} \and
    Bertie Vidgen\inst{2} \and
    Christian Schroeder de Witt\inst{1} \and
    Fabio Pizzati\inst{1} \and
    Katherine Elkins\inst{3} \and
    Supratik Mukhopadhyay\inst{4} \and
    Adel Bibi\inst{1} \and
    Aaron Purewal \inst{5} \and
    Botos Csaba\inst{1} \and
    Fabro Steibel\inst{6} \and
    Fazel Keshtkar \inst{7} \and
    Fazl Barez\inst{1} \and
    Genevieve Smith\inst{8} \and
    Gianluca Guadagni\inst{9} \and
    Jon Chun\inst{3} \and
    Jordi Cabot\inst{10,11} \and
    Joseph Marvin Imperial\inst{12,13} \and
    Juan A. Nolazco-Flores\inst{14} \and
    Lori Landay\inst{15} \and
    Matthew Jackson\inst{1} \and
    Philip H.S. Torr\inst{1} \and
    Trevor Darrell\inst{8} \and
    Yong Suk Lee\inst{16} \and
    Jakob Foerster\inst{1}
}

\institute{
    University of Oxford \and
    MLCommons \and
    Kenyon College \and
    Center for Computation \& Technology, Louisiana State University \and
    University of Texas at Austin \and
    Institute for Technology \& Society (ITS), Rio \and
    St. John's University \and
    University of California, Berkeley \and
    University of Virginia \and
    Luxembourg Institute of Science and Technology \and
    University of Luxembourg \and
    University of Bath \and
    National University Philippines \and
    ITESM \and
    Berklee College of Music \and
    University of Notre Dame
}

%
%

%
\authorrunning{\xspace}

\maketitle              
\setcounter{footnote}{0}

\begin{abstract}
Applications of Generative AI (Gen AI) are expected to revolutionize a number of different areas, ranging from science \& medicine to education. 
The potential for these seismic changes has triggered a lively debate about the potential risks of the technology, and resulted in calls for tighter regulation, in particular from some of the major tech companies who are leading in AI development. 
This regulation is likely to put at risk the budding field of open-source generative AI. 
Using a three-stage framework for Gen AI development (near, mid and long-term), we analyze the risks and opportunities of open-source generative AI models with similar capabilities to the ones currently available (near to mid-term) and with greater capabilities (long-term).
We argue that, overall, the benefits of open-source Gen AI outweigh its risks. As such, we encourage the open sourcing of models, training and evaluation data, and provide a set of recommendations and best practices for managing risks associated with open-source generative AI.\footnote{This work is an extension of \citet{eiras2024near} presented at ICML 2024.\\\textbf{Disclaimer}: This paper represents the collaborative effort of a diverse group of researchers, each bringing their own unique perspectives to the table. We note that not every viewpoint expressed within this work is necessarily unanimously agreed upon by all authors.}

\end{abstract}

\section{Introduction}

Generative AI (Gen AI), defined as \textit{``artificial intelligence that can generate novel content''} by conditioning its response on an input \citep{gozalo2023chatgpt} (e.g., large language or foundation models), is anticipated to profoundly impact a diverse array of domains including science \citep{ai4science2023impact}, the economy \citep{brynjolfsson2023generative}, education \citep{alahdab2023potential}, the environment \citep{rillig2023risks}, among many others. As a result, there has been significant socio-technical work undertaken to evaluate the broader risks and opportunities associated with these models, in a step towards a more nuanced and comprehensive understanding of their impacts \citep{bommasani2021opportunities}.

Parallel to these efforts is a debate on the \textit{openness of Gen AI} models. The digital economy heavily relies on open-source software, exemplified by over 60\% of global websites using open-source servers like Apache and Nginx \citep{lifshitz2021digital}. This prevalence is underscored by a 2021 European Union report, which concluded that `\textit{`overall, the [economic] benefits of open-source [software] greatly outweigh the costs associated with it''} \citep{blind2021impact}. Some developers of Gen AI models have chosen to openly release trained models (and sometimes data and code too), by leaning on this narrative and claiming that by doing so \textit{``[these models] can benefit everyone''} and that \textit{``it's safer [to release them]''} \citep{meta2023meta}. However, while there has been a flurry of reports and surveys on the impacts of general open-source software in areas such as innovation or research within the last few decades \citep{paulson2004empirical,schryen2009open,von2007open}, the discourse surrounding the openness of Gen AI models presents unique complexities due to the distinctive characteristics of this technology, including e.g., potential dual use and run-away technological progress.

\begin{wrapfigure}[22]{r}[12pt]{0.5\textwidth}
    \centering
    \vspace{-2.5em}
    \includegraphics[width=0.9\linewidth]{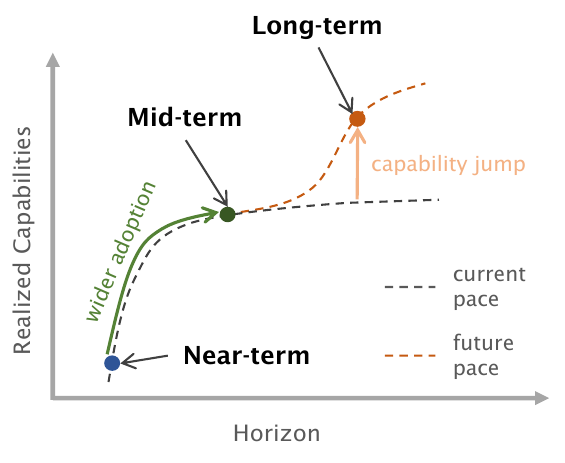}
    \vspace{-1em}
    \caption{\textbf{Three Development Stages for Generative AI Models}: \textit{near-term} is defined by early use and exploration of the technology in much of its current state; \textit{mid-term} is a result of the widespread adoption of the technology and further scaling at current pace; \textit{long-term} is the result of technological advances that enable greater AI capabilities.}
    \label{fig:development-stages}
\end{wrapfigure}

The urgency of assessing the risks and opportunities of open-source Gen AI is further underscored by recent regulatory developments around the world. The European Union (EU) Artificial Intelligence (AI) Act \citep{AIAct} has matured into the world's first comprehensive and enforceable regulatory framework on AI governance, and is set to introduce specific obligations to providers
of open-source general purpose AI models, and systems built thereon. Simultaneously, US President Joe Biden's Executive Order (EO) on AI~\citep{house_fact_2023} is thought to significantly affect open-source developers, while China's approach to AI regulation continues to be highly influenced by state intervention~\citep{china_cyber_2023, translate_interim_2023}. While these regulations may carve in stone certain aspects of future open-source Gen AI governance, fundamental questions surrounding concepts such as \textit{general-purpose AI models posing systemic risk} (EU AI Act) or \textit{dual-use foundation models} (Biden's EO) remain up to debate. Importantly, particularly in the case of the EU AI Act, many regulations have been designed to be adaptable in line with future technological progress. Our debate therefore remains highly relevant to open-source Gen AI governance.


\textbf{Contributions and Structure.} This paper argues that the success of open source in traditional software could be replicated in generative AI with well-defined and followed principles for responsible development and deployment. 

We start by defining different development/deployment stages of Gen AI models, dividing them into \textit{near}, \textit{mid} and \textit{long}-term stages based on adoption rates and technological advances (\S \ref{sec:prelim-stages}). To set up the discussion on the risks and opportunities, we present the current governance landscape with respect to open-source Gen AI (\S \ref{sec:prelim-governance}) and reexamine the definition of open source in the context of generative AI (\S \ref{sec:prelim-pipelines} and \ref{sec:taxonomy}). In \S \ref{sec:taxonomy}, we present an openness taxonomy for Gen AI by focusing on the individual components required to obtain models (\S \ref{sec:prelim-pipelines}), and a classification system with three broad classes (\textit{fully closed}, \textit{semi-open}, and \textit{fully open}) depending on their availability. We use this taxonomy to classify some of the most popular existing Large Language Models (LLMs), noting that (i) there is a \textit{notable skew towards closed source in model weights} (and even more in other components) (ii) \textit{open-source models currently underperform closed source ones}.\footnote{We focus on LLMs in these definitions and in \S \ref{sec:taxonomy} as this is the modality with the most prolific model development and open sourcing at the moment, but note that it would be easy to extend our analysis to other modalities.} An up-to-date version of the taxonomy of LLMs is also available on the following link: \href{https://open-source-llms.github.io}{https://open-source-llms.github.io}.

With this setup, we then focus on evaluating the risks and opportunities presented in the near to mid-term (\S \ref{sec:near-mid-term-impacts}) and long-term (\S \ref{sec:long-term-impacts}). 

Given the near to mid-term stage has already started, we start by taking a contrastive socio-technical analysis of the risks and benefits of Gen AI by considering 4 distinct areas of impact of the models (\S \ref{sec:near-mid-impact-contrastive}): \textbf{\RIDText{Research, Innovation and Development}}; \textbf{\SSText{Safety and Security}}; \textbf{\EAUText{Equity, Access and Usability}}; and \textbf{\BSAText{Broader Societal Aspects}}, and dividing the impacts into net positive or negative. For instance, we argue that open-source models can promote research and innovation through the empowerment of developers (\S \ref{sec:near-mid-term-empower-developers}) which we see as an overall positive impact, while the argument that open models cannot be rolled back or forced to update (\S \ref{sec:near-mid-term-roll-back}) is a net negative of these models. We then take a step back and contextualize often discussed risks from open sourcing generative AI models (\S \ref{sec:near-mid-impacts-real-risks}), critiquing common claims made about the process (\S \ref{sec:near-mid-impacts-claims}) and distinguishing between ``real'' and ``perceived'' risks given existing mitigation strategies (\S \ref{sec:near-mid-impacts-real-risks}). The combination of the significant benefits enumerated in \S \ref{sec:near-mid-impact-contrastive} and the risk clarification and mitigation strategies in \S \ref{sec:near-mid-impacts-real-risks} allow us to \textbf{strongly support the open sourcing of models in the near to mid-term stage}.

Our definition of the long-term stage assumes a technological advancement that enables greater AI capabilities (see \ref{sec:prelim-stages}). We start the discussion by defining what is commonly referred to as Artificial General Intelligence (AGI) and clarifying that the current debate on risks and opportunities of AGI is highly speculative. We then emphasize the importance of what we define as \textit{technical alignment}, and discuss how open sourcing AGI could help reduce existential risk associated with the technology (\S \ref{sec:long-term-x-risk}), as well as the benefits and non-existential risks that AGI poses (\S \ref{sec:long-term-benefits-non-x-risks}). Overall, we argue that open source (i) reduces the likelihood of existential risk by contributing to the development of technical alignment and maintaining the balance of power, as well as helps developing decentralized coordination mechanisms, and (ii) addresses some critical non-existential risks (e.g., cultural bias and social manipulation) while enhancing the benefits of AGI. These also form arguments to \textbf{support the open sourcing of models in the long-term stage}.

To balance the risks and opportunities presented in \S \ref{sec:near-mid-term-impacts} and \ref{sec:long-term-impacts}, we provide some recommendations for policy makers and developers as well as a set of best practices in \S \ref{sec:recommendations}. Overall, we strongly favor appropriate legislation and regulation of the improper \textbf{use} of Gen AI models, yet believe it is in society's best interest \textbf{not to restrict the development of open-source generative AI} by ensuring developers are not liable for the improper or illegal use of the resulting models (provided they are not developed to encourage such misuse).


\section{Preliminaries}
\label{sec:preliminaries}

To frame our analysis of the risks and benefits of responsibly open sourcing generative AI models, in this section we (i) outline a three-stage framework for AI development, (ii) present the current pipelines involved in training, evaluating and deploying Large Language Models (LLMs), and (iii) discuss the current state of open-source generative AI governance in various regions of the world. 

\subsection{Stages of Development of Generative AI Models}
\label{sec:prelim-stages}

Our three-part framework, presented visually in Figure \ref{fig:development-stages}, to describe the evolution of generative AI focuses on adoption rates and technological advancements instead of time elapsed (similar to \citeauthor{anthropic2023}, \citeyear{anthropic2023}). The \textbf{near-term} stage is defined by the early use and exploration of existing technology, such as deep learning with transformer and diffusion model architectures, utilizing large datasets. This phase is characterized by experimentation, with increasing levels of development, investment and adoption. The \textbf{mid-term} is defined by the widespread adoption and scaling of existing technology, and the exploitation of its benefits. We conceptualize this as moving along a predictable \textit{``capability curve''}, whereby more resources and usage will lead to greater benefits (and risks), but technological capabilities have not radically improved. Increasing use of multimodal models, agentic systems, and retrieval augmented generation are expected at this stage. The \textbf{long-term} is defined by a technological advance that will create dramatically greater AI capabilities, and therefore more risks and opportunities. This could manifest as a novel AI paradigm, a departure from traditional deep learning architectures, more efficient data utilization, among others, leading to more powerful AI models. In this paper, we focus primarily on analyzing the risks and opportunities of open-source Gen AI in the near to mid-term stages.

\subsection{Training, Evaluating, and Deploying Large Language Models}
\label{sec:prelim-pipelines}

\begin{figure}[t]
    \centering
    \includegraphics[width=\textwidth]{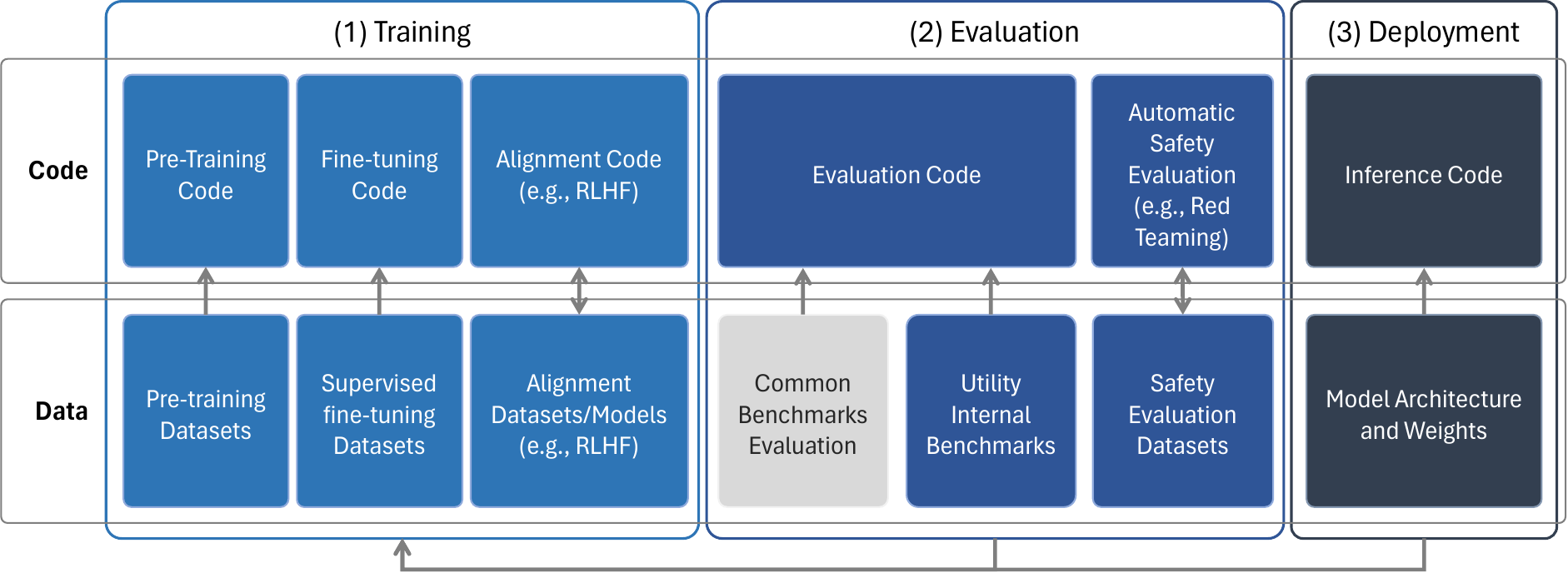}
    \caption{\textbf{Model Pipeline}: pipeline of model (1) training, (2) evaluation and (3) deployment analyzed in the report. The component Common Benchmarks Evaluation (in light gray) is included in the pipeline for completeness yet will not be analyzed in detail as these are commonly available and transversal to a substantial number of models. }
    \label{fig:openness-pipelines}
\end{figure}


The components typically involved in the (1) training, (2) evaluation, and (3) deployment of models are shown in Figure 1. The components can be divided into two categories: \textit{Code} and \textit{Data}.

Model training (part 1) processes can be grouped into three distinct stages:
\begin{enumerate}
    \item \textit{Pre-training}, where a model is exposed to large-scale datasets composed of trillions of tokens of data, typically scraped from the internet and usually uncurated. The goal is for the model to see a diversity of data, and through that process develop fundamental skills (e.g., grammar, vocabulary, text structure) and broad knowledge \citep{gao2020pile,radford2019language}. An example of a commonly used open-source dataset for pre-training LLMs such as LLaMA or GPT-J is The Pile which combines 22 smaller datasets into a diverse 825Gb text dataset \cite{gao2020pile,touvron2023llama,gpt-j}.
    \item \textit{Supervised fine-tuning (SFT)}, which is intended to correct for data quality issues in pre-training datasets. Usually, a much smaller amount of high quality data is used to improve model performance. Several works observe that at this stage the quality of the data used is essential to the downstream performance of the models \citep{zhou2024lima, ouyang2022training, touvron2023llama2, team2023gemini}, with the authors of LLaMA-2 pointing out that \textit{``by setting aside millions of examples from third-party datasets and using fewer but higher-quality examples from our own vendor-based annotation efforts, [their] results notably improved.''} \citep{touvron2023llama2}.
    \item \textit{Alignment}, which is used to create an application-specific version of the foundation model (e.g., a chatbot or translation model). Reinforcement Learning with Human Feedback (RLHF) or Direct Preference Optimisation (DPO) \citep{ouyang2022training, touvron2023llama2} is used to create a model that follows instructions and is better-aligned with human preferences. With RLHF, a dataset of human preferences over model outputs is used to train a Reward model, which in turn is used with a reinforcement learning algorithm (e.g., PPO; \citeauthor{schulman2017proximal}, \citeyear{schulman2017proximal}) to align the LLM. RLHF is not used in models released prior to 2022 \citep{brown2020language, xue2020mt5, smith2022using}, and it is unclear whether the RLHF is used in models such as PaLM-2 \citep{anil2023palm}.
\end{enumerate}

Once trained, models are usually evaluated (part 2) on openly available evaluation datasets such as MMLU or NaturalQuestions \citep{hendrycks2020measuring, kwiatkowski2019natural} as well as curated benchmarks such as HELM, BigBench EleutherAI’s Evaluation Harness \citep{liang2022holistic, srivastava2022beyond, eval-harness}. Some models are also evaluated on proprietary datasets held internally by developers, potentially by holding out some of the SFT/RLHF data from the training process \citep{touvron2023llama2}. However, there is little publicly available information on how this is implemented, and few details are shared about the composition of such datasets. On top of utility-based benchmarking, developers sometimes create safety evaluation mechanisms to proactively stress-test the outputs of the model. These include human-annotated safety evaluation datasets (e.g., through creating adversarial prompts), as well as automatic safety evaluation algorithms \citep{touvron2023llama2, yuan2023gpt4}. They are typically the result of applying techniques such as red teaming. Finally, at the deployment stage (part 3), content can be generated by running the inference code with the associated model weights.

\subsection{Open-source Gen AI Governance}
\label{sec:prelim-governance}

Recent years have seen the emergence of regulatory frameworks across the world that are already, or will soon, interact with the real-world governance of open-source Gen AI models. These efforts have been accompanied by increasing efforts at streamlining on the international stage, starting from 2023 G7 Hiroshima Summit and the Bletchley declaration~\citep{bletchley_2023}, and culminating in various national and transnational initiatives forming a network of AI safety institutes in the United Kingdom (UK), United States of America (US), European Union (EU), and elsewhere. Prior to the launch of ChatGPT on November 29th, 2023, such regulations were mostly targeted at (i) containing the spread of \textit{deepfakes} in order to safeguard election integrity -- e.g., the EU's 2022 amendments to the Digital Services Act --, or (ii) to exercise wider information control against the spread of ``rumors", such as the Chinese government's 2019 \textit{Regulations on the Administration of Online Audio and Video Information Services}~\citep{sheehan_chinas_2023}.
At the same time, the economic benefits of open-source AI models and systems have been almost unanimously recognized across the world. The launch of ChatGPT, and its rapid adoption among users worldwide, led policymakers to focus on general-purpose AI (GPAI) regulation.

\subsubsection{The EU AI Act}
The first \textit{comprehensive} regulatory framework governing general-purpose AI -- including provisions for open-source Gen AI -- may be the EU AI Act, which is expected to come into full force by 2026~\citep{AIAct}.
The legislation will apply to anyone putting AI services, or their outputs, on the EU market for professional purposes, while exempting recreational or academic use, as well as matters relevant to national security. It guards providers of open-source general-purpose models against risks emanating from downstream use by limiting the providers' responsibilities to a number of transparency obligations. These transparency obligations include the high-level documentation of training data provenance, as well a specification of intended use cases. 
Entities deploying Gen AI \textit{deepfakes} are required to disclose their AI-generated nature. These requirements will apply to small business owners to a lesser degree. While comprehensive, the EU AI Act will not apply to recreational or research use and will be superseded by the EU member states' individual national security interests. Open-source Gen AI providers may face additional procedures and obligations if their models are classified as \textit{general-purpose AI (GPAI) models of systemic risk}, an intentionally vaguely defined criterion that will be adapted as technology progresses. Importantly, the EU AI Act, as perhaps the EU's first transnational legislation, explicitly affirms the economic benefits of open-source AI.

\subsubsection{Biden's Executive Order}
President Biden's 2023 \textit{Executive Order (EO) on Safe, Secure, and Trustworthy Artificial Intelligence}~\citep{house_fact_2023} continues to follow a \textit{``soft law''} approach of earlier EOs, largely trading enforceable regulation for voluntary industry commitments~\citep{pricewaterhousecoopers_overview_2024}. Safety and security measures surrounding AI technology include requirements for developers to share red-teaming results with the US federal government, and for companies working on ``dual-use'' foundation models (\textit{i.e.}, systems with civilian and military applications) and/or with large compute clusters to provide regular activity reports. The National Institute of Standards and Technology (NIST) is set up to play a key role in developing standards for secure and safe AI. Instead of placing hard restrictions on the use of certain AI technology (as the EU AI Act explicitly does), Biden's EO focuses on promoting best practices, evaluations, and standard development across a wide variety of aspects including security and risk mitigation. For example, it includes references to biological weapons, AI-generated content watermarking, and labor market impacts, and, additionally, measures for attracting foreign national AI talent through streamlining visa procedures and by providing assistance to small businesses and developers. 
National security interests are also formulated, including the reporting of foreign users of US Infrastructure as a Service (IaaS) products, as well as promoting the development of AI-driven tools to detect cyber vulnerabilities.

\subsubsection{China's Gen AI Legislation} 
The earliest legal framework specifically targeting Gen AI models and systems, the Chinese government's \textit{Provisional Administrative Measures of Generative Artificial Intelligence Services (Generative AI Measures)} \citep{china_cyber_2023,translate_interim_2023}, came into force in China in August 2023. These regulations pose strict obligations on providers of Gen AI, ranging from outcome-driven provisions (e.g., requiring generative AI services to not produce illegal or untruthful content) to provenance obligations on training data and model weights, and measures targeted to protect intellectual property and privacy rights \citep{chong_china_2023}. From the point of view of open-source model developers, the inability to predict future downstream use of models and systems provided introduces legal risks that require regulatory containment. Although open-source Gen AI plays a significant role in the Chinese economy, however, these regulations do not seem to target open-source (GP)AI models specifically \citep{china_emerging_2024}. 

\subsubsection{The Middle East}
\textbf{Saudi Arabia.} 
In August 2019, as part of Saudi Arabia's Vision 2030 introduced by Crown Prince Mohammed Bin Salman, the Saudi Data and AI Authority (SDAIA) was established by a royal decree. SDAIA aims to advance this vision, with the National Center for AI serving as a key component.
Saudi Arabia, through SDAIA, has adapted and released its first version of AI ethics in September 2023 \citep{SDAIA_AI_ethics}. The document outlines Saudi's stance on AI risks, categorized from minimal to unacceptable risks with a comprehensive risk management plan covering data, algorithms, compliance, operations, legality, and regulatory risks. The AI ethics strongly supports the transparent development and deployment of AI
, reflecting that \textit{``transparent and explainable algorithms ensure that stakeholders affected by AI systems [...] are fully informed when an outcome is processed by the AI''}. Moreover, SDAIA has quickly embraced the generative AI wave. In collaboration with NVIDIA, SDAIA developed ``Allam'' \citep{Allam}, Saudi Arabia's first national-level LLM model, an Arabic LLM designed to provide summaries and answer questions, drawing information from cross-checked online sources. While Allam was closed source 
and only a beta version interface is accessible, there are still several pieces of evidences that Saudi Arabia is in favor of open-source. For instance, the Digital Government Authority \citep{digital_gov_auth} issued free and open-source government software licenses to 6 government agencies in 2022. This entails reviewing and publishing the source code ``in a way that opens the field of cooperation and unified standards among government agencies". The general directions with the laid down compliance regulations, stated principles, and open source government suggest that Saudi Arabia is in favor of open source.

\textbf{United Arab Emirates (UAE).} In October 2017, the UAE Government launched the \textit{``UAE Artificial Intelligence Strategy''} \citep{UAE_strategy_Gov}, spanning sectors from education to space. Shortly after, His Excellency, Omar Al Olama became the world's first AI minister. The UAE has been in favor of open-source in their policies, for instance, as stated in the strategy \textit{``Objective 7: Provide the data and the supporting infrastructure essential to become a test bed for AI''} and that \textit{``The UAE has an opportunity to become a leader in available open data for training and developing AI systems''}. Moreover, the strategy states that \textit{``The UAE’s ambition is to create a data-sharing program, providing shared open and standardized AI-ready data, collected through a consistent data standard''}. More recently, the UAE through the Technology Innovation Institute (TII) has open-sourced its LLM Falcon \citep{falconllm}, including its 180B parameter version, for both research and commercial use \citep{falconllm_opensource}. This all indicates the UAE’s positive take towards open-source models.

\subsubsection{AI Regulation Efforts in Other Countries}

In 2019, the Organization for Economic Co-operation and Development (OECD) introduced their AI Principles, a recommendation by the council on general-purpose AI. These principles were ratified by the G20 council, and have been adopted by at least 42 of the organization's participating countries \citep{noauthor_oecds_nodate, australian_2024}. 

Some countries have on-going legislation efforts or have issued policies specifically on Gen AI, addressing mainly sector-based issues. These include Australia \citep{noauthor_interim_nodate}, Canada \citep{noauthor_guidelines_nodate}, New Zealand \citep{kaldestad_new_2023}, Norway \citep{council2023ghost}, Singapore \citep{noauthor_mas_nodate, noauthor_first_nodate}, among others. India has published working papers on the issue of AI and enacted the Digital Personal Data Protection Act in 2023 tackling privacy issues related to Gen AI \citep{india_regulation} -- it is yet to regulate on general-purpose Gen AI and the open sourcing of models. Brazil is working on two main legislative proposals to regulate AI, one inspired in the US framework (Bill no. 21, from 2021) and another inspired on the EU framework (Bill No. 2338, from 2023), yet these do not have provisions for open-source Gen AI models. A few other countries are in the process of running public consultations on how to regulate generative AI, such as the case of Chile \citep{httpsmagnetcl_articulo_nodate} and Uruguay \citep{noauthor_mesa_nodate}.

\section{Openness Taxonomy of Generative AI Models}
\label{sec:taxonomy}


Model developers decide whether to make each component of the training, evaluation and deployment pipeline (Figure \ref{fig:openness-pipelines}) \textit{private} or \textit{public}, with varying levels of restrictions for the latter. 
For instance, the developers of LLaMA-2 have publicly released the model architecture and weights, yet they have not shared the code or reward model for Reinforcement Learning from Human Feedback (RLHF) used in the Alignment components \citep{touvron2023llama}. 
To properly evaluate the openness of each component, we introduce a classification scale in \S \ref{sec:openness_classification}, which we then apply to 45 high impact LLMs in \S \ref{sec:taxonomy-llms}. This will help contextualizing the risks and opportunities discussed in \S \ref{sec:near-mid-term-impacts} and \S \ref{sec:long-term-impacts}. An up-to-date version of the taxonomy of LLMs is also available on the following link: \href{https://open-source-llms.github.io}{https://open-source-llms.github.io}.

\subsection{Classifying Openness for Generative AI Code and Data}
\label{sec:openness_classification}

\begin{figure}[t]
    \centering
    \includegraphics[width=\textwidth]{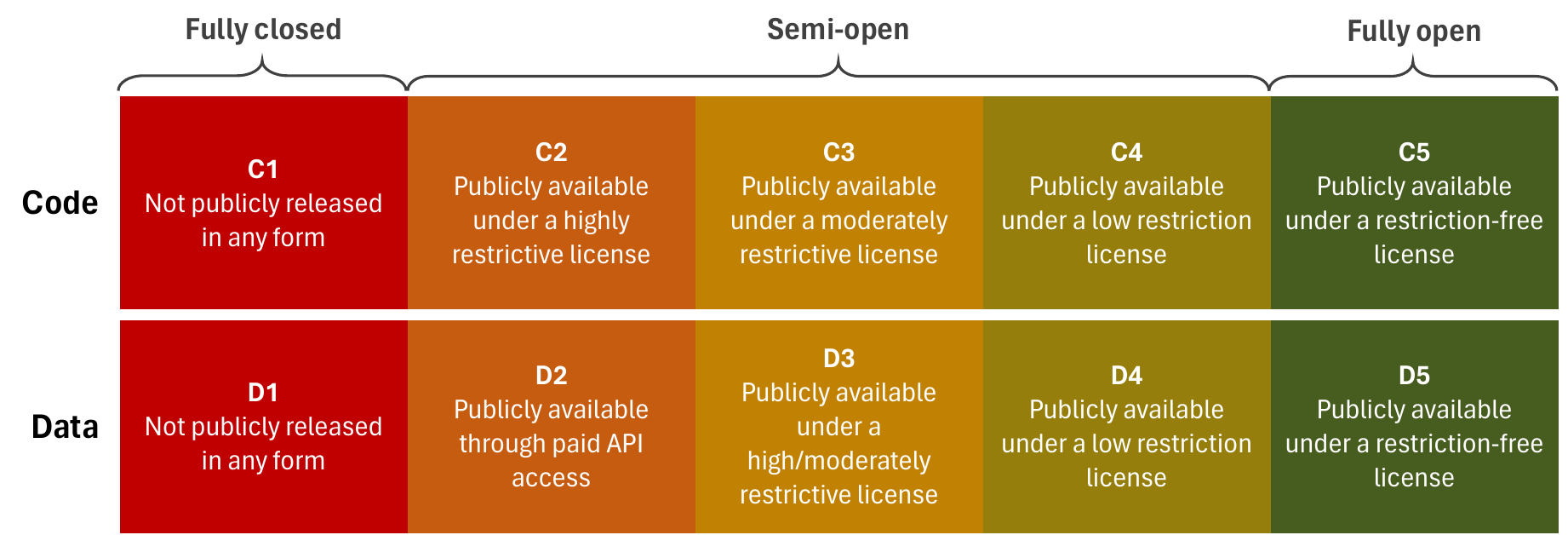}
    \caption{\textbf{Openness Scale}: categorization of the levels of openness of the code and data of each model component. See Table \ref{tab:licenses} for a reference on the restrictions imposed by each license.}
    \label{fig:openness-levels}
\end{figure}



We introduce a framework for categorizing the openness of each component of the pipeline in Figure \ref{fig:openness-pipelines}. At the highest level, a \textbf{fully closed} component is not publicly accessible in any form \citep{rae2022scaling}. In contrast, a \textbf{semi-open} component is publicly accessible but with certain limitations on access or use, or it is available in a restricted manner, such as through an Application Programming Interface (API) \citep{achiam2023gpt}. Finally, a \textbf{fully open} component is available to the public without any restrictions on its use \citep{xu2022systematic}. Further, the semi-open category comprises three subcategories, delineating varied openness levels as detailed in Figure \ref{fig:openness-levels}. Distinctions are made between Code (\COneText --\CFiveText) and Data (\DOneText --\DFiveText) components, where \CFiveText/\DFiveText represents unrestricted availability and \COneText/\DOneText denotes complete unavailability. 

Note that, in some cases, some of the code or data for a component will be publicly available while other portions of it will be kept private. In such cases where the private code/data is key to reproducing the pipeline accurately, we will categorize this as a fully-closed component; otherwise the component’s categorization will depend on the license provided for the public code/data available.




To evaluate the licenses we introduce a point-based system where each license gets 1 point (for a total maximum of 5) for allowing each of the following: \textit{can use a component for research purposes} (\textbf{Research}), \textit{can use a component for any commercial purposes} (\textbf{Commercial Purposes}), \textit{can modify a component as desired (with notice)} (\textbf{Modify as Desired}), \textit{can copyright derivative} (\textbf{Copyright Derivative Work}), \textit{publicly shared derivative work can use another license} (\textbf{Other license derivative work}). 

The total number of points is indicative of a license's restrictiveness. A \textbf{Highly restrictive} license scores 0-1 points, aligning with openness levels of code C2 and data D3, imposing significant limitations. A \textbf{Moderately restrictive} license, scoring 2-3 points (code C3 and data D3), allows more flexibility but with some limitations. Licenses scoring 4 points are \textbf{Slightly restrictive} (code C4 and data D4), offering broader usage rights with minimal restrictions. Finally, a \textbf{Restriction free} license scores 5 points, indicating the highest level of openness (code C5 and data D5), permitting all forms of use, modification, and distribution without constraints.

\begin{table}[t]
    \begin{adjustwidth}{-1cm}{-1cm}
    {
    \scriptsize
    \centering
    \begin{tabular}{>{\hspace{0pt}}m{0.14\linewidth}>{\centering\hspace{0pt}}m{0.07\linewidth}>{\centering\hspace{0pt}}m{0.112\linewidth}>{\centering\hspace{0pt}}m{0.094\linewidth}>{\centering\hspace{0pt}}m{0.121\linewidth}>{\centering\hspace{0pt}}m{0.127\linewidth}>{\centering\hspace{0pt}}m{0.113\linewidth}>{\centering\hspace{0pt}}m{0.09\linewidth}>{\centering\arraybackslash\hspace{0pt}}m{0.083\linewidth}}
    \toprule
    \textbf{License}        & \textbf{Research} & \textbf{Commercial Purposes} & \textbf{Modify as Desired} & \textbf{Copyright derivative work} & \textbf{Other license for derivative} & \textbf{Final score}            & \textbf{Code Openness} & \textbf{Data Openness}  \\ \midrulenospacing
    MIT/Mod. MIT            & Y                 & Y                            & Y                          & Y                                  & Y                                     & 5\par{}(Restriction free)       & \CFiveTable                     & \DFiveTable                      \\ \hline
    Apache 2.0              & Y                 & Y                            & Y                          & Y                                  & Y                                     & 5\par{}(Restriction free)       & \CFiveTable                     & \DFiveTable                      \\ \hline
    Common Crawl\par{}(ComCrawl) & Y                 & Y                            & Y                          & Y                                  & Y                                     & 5\par{}(Restriction free)       & \CFiveTable                     & \DFiveTable                      \\ \hline
    BSD-3                   & Y                 & Y                            & Y                          & Y                                  & Y                                     & 5\par{}(Restriction free)       & \CFiveTable                     & \DFiveTable                      \\ \hline
    RAIL                    & Y                 & Y                            & Y                          & Y                                  & N                                     & 4\par{}(Slightly restrictive)   & \CFourTable                     & \DFourTable                      \\ \hline
    LLaMA-2/3                 & Y                 & Y$^2$                            & N                          & Y                                  & N                                     & 3\par{}(Moderately restrictive) & \CThreeTable                     & \DThreeTable                      \\ \hline
    DBRX                 & Y                 & Y$^2$                            & N                          & Y                                  & N                                     & 3\par{}(Moderately restrictive) & \CThreeTable                     & \DThreeTable                      \\ \hline
    C4AI                 & Y                 & N                            & Y                          & Y                                  & N                                     & 3\par{}(Moderately restrictive) & \CThreeTable                     & \DThreeTable                      \\ \hline
    ODC-By                  & Y                 & Y                            & Y                          & Y                                  & N                                     & 4\par{}(Slightly restrictive)   & \NATable                    & \DFourTable                      \\ \hline
    CodeT5 Data             & Y                 & Y                            & Y                          & Y                                  & N                                     & 4\par{}(Slightly restrictive)   & \NATable                    & \DFourTable                      \\ \hline
    RedPajama Data\par{}(Full)   & Y                 & Y                            & Y                          & Y                                  & N                                     & 4\par{}(Slightly restrictive)   & \NATable                    & \DFourTable                      \\ \hline
    OPT Data                & Y                 & N                            & N                          & N                                  & N                                     & 1\par{}(Highly restrictive)     & \NATable                    & \DThreeTable                      \\ \hline
    GLM-130B Data           & Y                 & N                            & N                          & N                                  & N                                     & 1\par{}(Highly restrictive)     & \NATable                    & \DThreeTable                      \\ \hline
    Falcon-180B Data        & Y                 & Y                            & Y                          & Y                                  & Y                                     & 5\par{}(Restriction free)       & \NATable                    & \DFiveTable               \\ \bottomrulenospacing
    \end{tabular}
    }
    \vspace{0.5em}
    \caption{\textbf{License Openness Taxonomy}: categorization of commonly used licenses in a variety of relevant open source criteria, and resulting code and data openness categories. $^2$For models with up to 700M monthly active users.}
    \label{tab:licenses}
    \end{adjustwidth}
\end{table}

In Table \ref{tab:licenses} we show how each of the licenses used in the pipelines of current models ranks in terms of the point system, and consequently how they fit within the openness scale. 

\subsection{Openness Taxonomy of Current Large Language Models}
\label{sec:taxonomy-llms}

In this section we analyzed the pipeline components of 45 high-impact LLMs released from 2019 to 2024, chosen by optimizing three key impact metrics:
\begin{itemize}
    \item \textbf{ChatBot Arena Elo Rating}: a crowdsourced benchmarking score obtained by pitting models against each other (released in May 2023, so older models will be underrepresented).
    \item \textbf{Google Scholar Citations}: a measure of the academic impact of each of the models.
    \item \textbf{HuggingFace Downloads Last Month}: a measure of the use of each of the models that has been openly released on HuggingFace.
\end{itemize}

\renewcommand{\arraystretch}{1.4}
\begin{table}
    \begin{adjustwidth}{-2cm}{-2cm}
    {
    \scriptsize
    \centering

\begin{tabular}{>{\hspace{0pt}}m{0.10\linewidth}>{\centering\hspace{0pt}}m{0.14\linewidth}>{\centering\hspace{0pt}}m{0.11\linewidth}>{\centering\hspace{0pt}}m{0.09\linewidth}>{\centering\hspace{0pt}}m{0.13\linewidth}>{\centering\hspace{0pt}}m{0.13\linewidth}>{\centering\hspace{0pt}}m{0.15\linewidth}>{\hspace{0pt}}m{0.11\linewidth}}

    \toprule
    \multirow{2}[6]{\linewidth}{\hspace{0pt}\textbf{Model}} & \multirow{2}[6]{\linewidth}{\hspace{0pt}\Centering{}\textbf{Developer}} & \multirow{2}[6]{\linewidth}{\hspace{0pt}\Centering{}\textbf{Largest Model Size (params)}} & \multirow{2}[6]{\linewidth}{\hspace{0pt}\Centering{}\textbf{Release Date}} & \multicolumn{3}{>{\Centering\hspace{0pt}}m{0.41\linewidth}}{\textbf{Impact Metrics}}                             & \multirow{2}[6]{\linewidth}{\hspace{0pt}\textbf{Reference(s)}}                                                                             \\ \cmidrule{5-7}
                                                              &                                                                           &                                                                                            &                                                                              & \textbf{ChatBot Arena Elo Rating} & \textbf{Google Scholar Citations} & \textbf{HuggingFace Downloads Last Month} &                                                                                                   \\
\end{tabular}
    }
    \vspace{0.5em}
    \caption{\textbf{Model Information}: table containing the basic information about each of the models classified under the openness taxonomy. Developers highlighted in \colorbox{CompanyColor}{purple} correspond to companies, in \colorbox{NonProfitColor}{pink} are non-profit entities, and in \colorbox{GovernmentColor}{light blue} are government institutes. All data accessed on 8th of May 2024.}
    \label{tab:model_list}
    \end{adjustwidth}
\end{table}

While we included models that scored high on any of these metrics, we also decided to include other released models for the sake of diversity. Table \ref{tab:model_list} lists the characteristics of the models analyzed.

\begin{figure}[ht]
    \centering
    \begin{subfigure}[b]{0.48\textwidth}
        \centering
        \includegraphics[width=\linewidth]{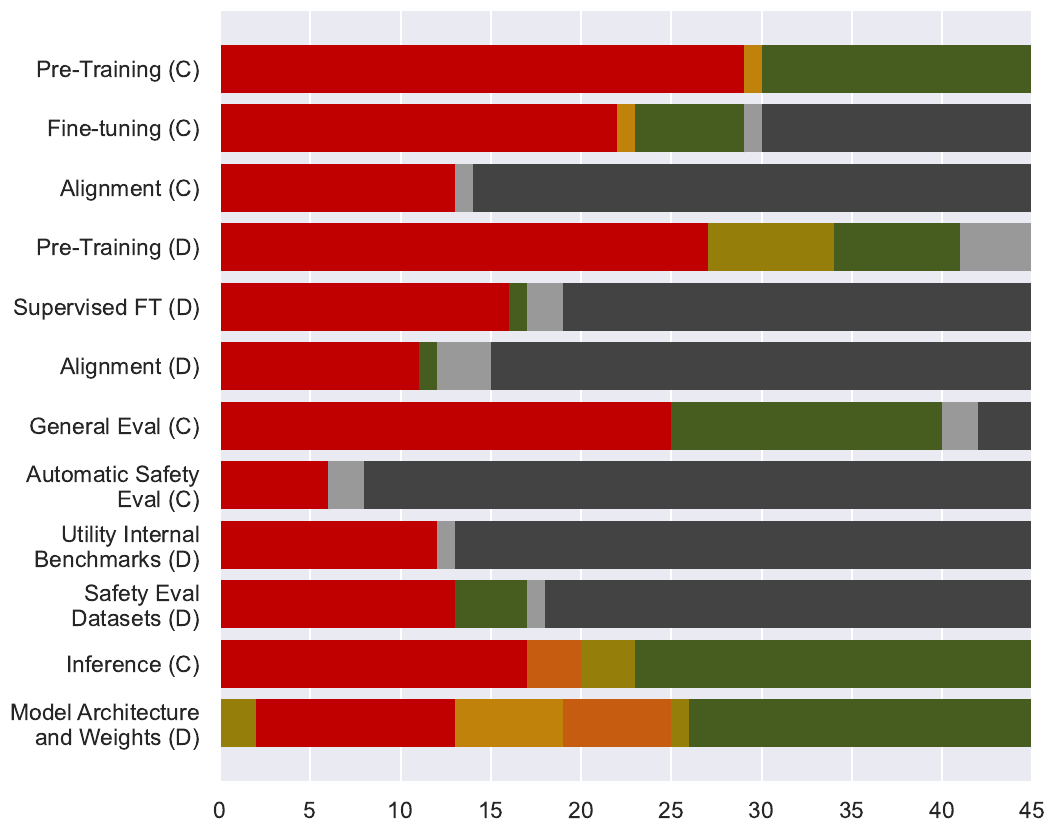}
        \caption{Openness Levels by Pipeline Component}
        \label{fig:classification_distribution}
    \end{subfigure}
    \hfill 
    \begin{subfigure}[b]{0.5\textwidth}
        \centering
        \includegraphics[width=\linewidth]{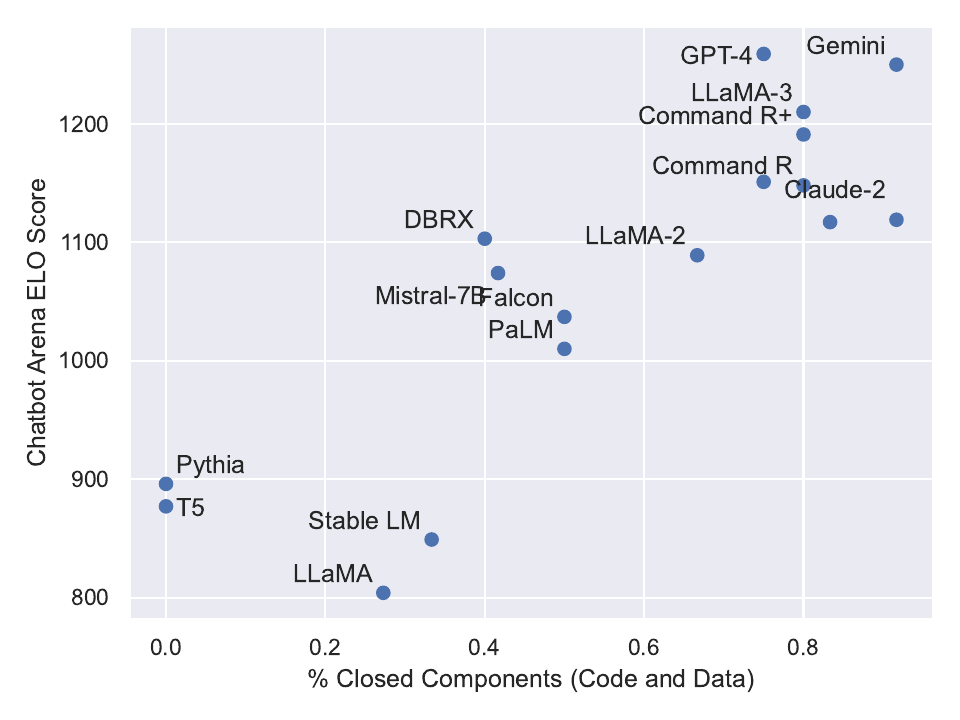}
        \caption{Closed Models Currently Outperform Open Ones}
        \label{fig:arena_elo_scores_vs_closed_components}
    \end{subfigure}
    \caption{\textbf{Taxonomy Analysis}: (a) shows the openness level distribution for each of the pipeline components of the 45 LLMs studied. Color legend: \colorbox{LOne}{\color{white} C1/D1}, \colorbox{LTwo}{\color{white} C2/D2}, \colorbox{LThree}{\color{white} C3/D3}, \colorbox{LFour}{\color{white} C4/D4}, \colorbox{LFive}{\color{white} C5/D5}, \colorbox{LUnknown}{\color{white} ?} (unknown or not publicly available), \colorbox{LNA}{\color{white} N/A} (not applicable). For conciseness, we use "FT" as a stand in for "Fine-Tuning". (b) plots the percentage of closed components for the studied models and their Chatbot Arena ELO Score.}
\end{figure}

Table \ref{tab:classification} shows how each model component ranks in terms of the openness scale. There are a couple of observations we can take from summarizing some details of Table \ref{tab:model_list} and \ref{tab:classification}: 
\begin{itemize}
    \item[\textbf{O1.}] \textbf{Providers Open Source Weights, Not Data or Evaluation Code.} Figure \ref{fig:classification_distribution} shows the distribution of openness levels for each of the pipeline components analyzed which clearly highlights a balance between open and closed source deployed components (inference code and weights). However, \textit{a notable skew exists towards closed source in training data (such as fine-tuning and alignment) and, importantly, in safety evaluation code and data}. As discussed in the next sections, for open-source's advantages to be fully leveraged and risks mitigated, a significant shift in this landscape is necessary, achievable only through responsible open-source generative AI development and deployment.
    \item[\textbf{O2.}] \textbf{Open-Source Models Underperform Closed Sourced Ones.} Figure \ref{fig:arena_elo_scores_vs_closed_components} plots the Chatbot Arena ELO Score against the \% of Closed Components (Code and Data) for the models in which the score is available. It is clear to observe that the \textit{most open-source} models such as Pythia or T5 are the \textit{worst performing}, indicating that i) providers are open sourcing a smaller amount of their component pipelines, and ii) closed source models outperform open-source ones.
\end{itemize}



\textbf{Important disclaimer}: the information in Table \ref{tab:classification} pertains only to the openness of each of the components of each model’s pipeline, and does not contain information regarding the reproducibility from the references available for that model. For example, the accompanying paper for GLM-130B \citep{zeng2022glm} and Falcon \citep{almazrouei2023falcon} includes a significant amount of detail regarding the training procedure, which would likely simplify the process of reproducing their pre-training results, whereas the GPT-4 \citep{yuan2023gpt4} or Gemini \citep{team2023gemini} reports do not include those details. However, from the perspective of our taxonomy all of these models’ pre-training code is classified as C1 as none of the developers have released it. There is an important distinction between disclosing certain reproducibility details or none at each stage of the pipeline. However, assessing the extent and quality of this information exceeds the purview of this report, necessitating a comprehensive reproducibility investigation.

\newgeometry{
    margin=0.9cm,
    noheadfoot, nomarginpar,
    footskip=1.25em,
}
\begin{landscape}
{
    \scriptsize
\begin{longtable}{>{\hspace{0pt}}m{0.072\linewidth}|>{\centering\hspace{0pt}}m{0.06\linewidth}>{\centering\hspace{0pt}}m{0.06\linewidth}>{\centering\hspace{0pt}}m{0.071\linewidth}|>{\centering\hspace{0pt}}m{0.063\linewidth}>{\centering\hspace{0pt}}m{0.069\linewidth}>{\centering\hspace{0pt}}m{0.071\linewidth}|>{\centering\hspace{0pt}}m{0.065\linewidth}>{\centering\hspace{0pt}}m{0.075\linewidth}|>{\centering\hspace{0pt}}m{0.073\linewidth}>{\centering\hspace{0pt}}m{0.073\linewidth}|>{\centering\hspace{0pt}}m{0.06\linewidth}|>{\centering\arraybackslash\hspace{0pt}}m{0.119\linewidth}}
    
    \toprule
    \multirow{3}[6]{0.072\linewidth}{\hspace{0pt}\textbf{Model}} & \multicolumn{6}{>{\centering\hspace{0pt}}m{0.394\linewidth}}{\textbf{(1) Training}}                                                                                                                   & \multicolumn{4}{>{\centering\hspace{0pt}}m{0.286\linewidth}}{\textbf{(2) Evaluation}}                                                                      & \multicolumn{2}{>{\centering\arraybackslash\hspace{0pt}}m{0.179\linewidth}}{\textbf{(3) Deployment}}  \\ \cmidrule{2-13}
    
    & \multicolumn{3}{>{\centering\hspace{0pt}}m{0.191\linewidth}}{\textbf{Code}}       & \multicolumn{3}{>{\centering\hspace{0pt}}m{0.203\linewidth}}{\textbf{Data }}                                      & \multicolumn{2}{>{\centering\hspace{0pt}}m{0.14\linewidth}}{\textbf{Code }} & \multicolumn{2}{>{\centering\hspace{0pt}}m{0.146\linewidth}}{\textbf{Data }} & \textbf{Code}          & \textbf{Data}                                                                \\ \cmidrule{2-13} 
    
    & \textbf{Pre-Training}  & \textbf{Fine-tuning}   & \textbf{Alignment (e.g., RLHF)} & \textbf{Pre-Training}                         & \textbf{Supervised Fine-Tuning} & \textbf{Alignment (e.g., RLHF)} & \textbf{General Evaluation}  & \textbf{Automatic Safety Evaluation}         & \textbf{Utility Internal Benchmarks} & \textbf{Safety Evaluation Datasets}   & \textbf{Inference}     & \textbf{Model Architecture and Weights}                                      \endfirsthead
    \midrulenospacing
    GPT-2                                                     & \COneTable                   & \NATable                 & \NATable                          & \DOneTable                                           & \NATable                          & \NATable                          & \COneTable                         & \NATable                                       & \DOneTable                                  & \NATable                                & \CFiveTable\par{}(Mod. MIT)     & \DFiveTable\par{}(Mod. MIT)                                                           \\\hline
T5                                                        & \CFiveTable\par{}(Apache 2.0)   & \CFiveTable\par{}(Apache 2.0)   & \NATable                          & \DFourTable\par{}(ODC-By)                              & \NATable                          & \NATable                          & \CFiveTable\par{}(Apache 2.0)         & \NATable                                       & \NATable                               & \NATable                                & \CFiveTable\par{}(Apache 2.0)   & \DFiveTable\par{}(Apache 2.0)                                                         \\\hline
GPT-3                                                     & \COneTable                   & \COneTable                   & \NATable                          & \DOneTable                                           & \NATable                          & \NATable                          & \COneTable                         & \NATable                                       & \DOneTable                                  & \NATable                                & \COneTable                   & \DTwoTable                                                                           \\
\end{longtable}
}
\end{landscape}
\restoregeometry

\section{Near to Mid-Term Impacts of Open-Source Generative AI}
\label{sec:near-mid-term-impacts}

\begin{figure}[t]
    \centering
    \includegraphics[width=\textwidth]{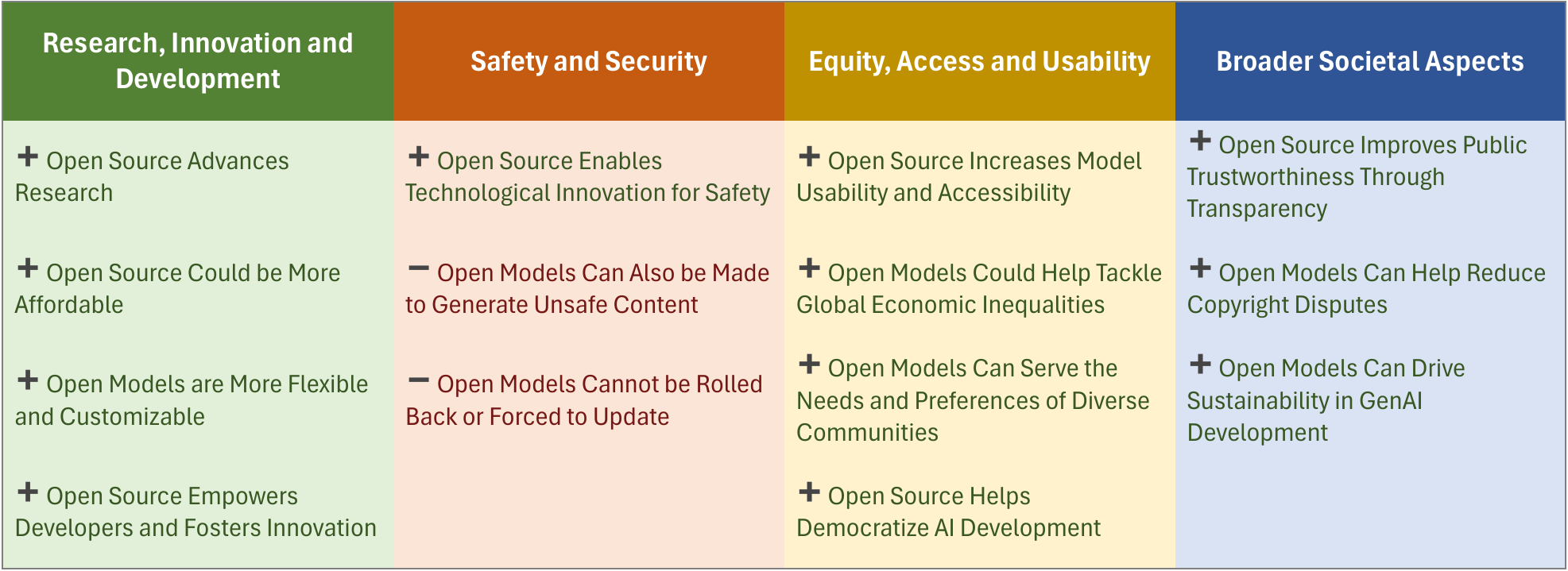}
    \caption{\textbf{Near to Mid-term Impacts of Open-Source Models}: specific impacts of open-source Gen AI models categorized by area of impact and whether they are positive (+) or negative (-).}
    \label{fig:near-mid-term-areas}
\end{figure}

In this section, we focus on discussing the risks and opportunities that arise from open sourcing Generative AI models in the near to mid-term horizon. We start by taking a direct, socio-technical contrastive approach, analyzing impacts of standalone open-source models compared to closed ones in several key areas (\S \ref{sec:near-mid-impact-contrastive}). Building on this foundation, in \S \ref{sec:near-mid-impacts-real-risks} we take a holistic approach to contextualizing the relative risks of open-source vs risks of close-source Generative AI models. This analysis considers risk in the context of existing baseline risks of other technologies, as well as the broader scope of societal governance and risk mitigation strategies.

Note that, as per our definition in the preliminaries (\S \ref{sec:preliminaries}), the near to mid-term phase has already started, corresponding to an early experimentation with existing models, and is book-ended by the widespread adoption and scaling of the available technology. Importantly, it excludes the possibility of a technological advancement that leads to a dramatic capability change in these systems, which is instead captured in the long-term phase of our framework. Unless explicitly stated, ``open source'' refers to all artifacts and components of the system including model, weights and datasets.

\subsection{Contrastive Socio-Technical Analysis of Risks and Benefits}
\label{sec:near-mid-impact-contrastive}

Our focus is on how open-source Generative AI catalyzes, minimizes or creates risks and opportunities compared to closed source models. To analyze these, we consider 4 different areas of impact for these models:
\begin{itemize}
    \item \textbf{\RIDText{Research, Innovation and Development}}
    \item \textbf{\SSText{Safety and Security}}
    \item \textbf{\EAUText{Equity, Access and Usability}}
    \item \textbf{\BSAText{Broader Societal Aspects}}
\end{itemize}

For each of these areas, we introduce a set of potential benefits or risks, first by contextualizing each issue and then discussing the \textit{marginal} advantage or drawback that open-source models might have over closed source ones. These are summarized in Figure \ref{fig:near-mid-term-areas}. Before we dive into each of these areas, we discuss the challenges of assessing impacts.

\textbf{The Challenges of Assessing Risks and Benefits.} Generative AI systems can be evaluated through a variety of currently available methods and frameworks, such as benchmarks like HELM and Big-Bench for task evaluation, Chatbot Arena for crowd-sourced model comparisons, and red teaming for exploratory evaluation \cite{guo2023evaluating, liang2023holistic, srivastava2023imitation}. However, these approaches face limitations like limited ecological validity and data contamination \cite{li2023static, sainz2023nlp, zhou2023dont}, and provide only a partial view of how models will perform in real-world settings. 
In response, some experts suggest socio-technical evaluations that are focused on real-world applications \cite{weidinger2023,solaiman2023evaluating}. This is supported by calls for comprehensive pre-release audits of models, datasets, and research artifacts \cite{derczynski2023assessing, M_kander_2023, 10.1145/3600211.3604712}. 
However, even holistic approaches to evaluation face substantial challenges, such as the rapid and unpredictable evolution of AI capabilities, the difficulty of standardizing measurements due to the fast pace of change, and the research community's limited insight into AI's industrial applications. This invariably leads to partial and incomplete evidence.
Considering these aspects, in this section we use a variety of evidence to critically examine and support our arguments. Nonetheless, it is important to recognize the inherent challenges in reaching definitive conclusions, underscoring the need for readers to recognize the evidence's limitations and for the community to improve its relevance and quality.

\subsubsection{\RIDText{Research, Innovation and Development}}

Research, innovation, and development are pivotal to the economic and technological progress of organizations and nations. Generative AI represents a transformative force in these domains, offering threefold benefits: it accelerates scientific breakthroughs through enhanced hypothesis generation, data analytics, and complex system simulations; it drives the creation of innovative products, services, and methodologies; and it refines operational efficiencies by streamlining workflows and optimizing resource management. Within this context, open-source Generative AI presents substantial opportunities to outperform closed-source models. We present four such benefits below.

\paragraph{\positiveImpact{Open Source Advances Research}}
\label{sec:near-mid-term-impacts-research}

Diverse and innovative research is crucial for advancing machine learning and understanding AI's societal impact. It introduces varied perspectives and novel solutions, improving AI's efficiency, effectiveness, and ethics. Historically, restricted access to modeling techniques and closed source AI systems has limited this diversity and the reproducibility of the outputs. Transparent, accessible, and reproducible research is essential for scientific progress and aligning AI development with ethical and societal norms.

Open source has emerged as a powerful solution to these challenges, providing unprecedented access to Large Language Models (LLMs) and other Generative AI artifacts. Particularly, open sourcing model pipelines enables (1) the inspection and understanding of the models, (2) the reproducibility of existing research and (3) the fostering of new advances in the Generative AI field. 

Firstly, with open-source artifacts researchers are equipped to perform white-box mechanistic probing, such as causal reasoning and inference analysis \citep{jin2023can, jin2023cladder}, and black-box probing through prompt engineering, including techniques like jailbreaking attacks and prompt injections \citep{liu2023trustworthy}. These approaches are invaluable for uncovering unsafe, harmful, and biased content within AI models, as detailed in \S \ref{sec:near-mid-term-impacts-safety-innovation}. Additionally, open source facilitates the investigation into how LLMs memorize copyrighted data, a critical concern that could lead to lawsuits if not properly addressed (see \S \ref{sec:near-mid-term-impacts-broader-societal}). Secondly, open-source methods, datasets and trained models strengthen the trust in research outputs through reproducibility. This streamlines the process of navigating through vast volumes of published papers, facilitating quicker turnaround times for academic contributions and innovations that build upon existing impactful research \citep{spirling2023open}. Finally, open models have also been instrumental in fostering breakthroughs, such as Direct Preference Optimization (DPO) \citep{rafailov2023direct}, providing cost-efficient alternatives to Reinforcement Learning from Human Feedback (RLHF) \citep{ouyang2022training}. These innovations demonstrate that open models can achieve performances comparable to their closed counterparts on select tasks. 

Conversely, closed models often only grant limited access through API calls, and may restrict access to essential model generation outputs such as logits and token probabilities. Such limitations bar researchers from forming deeper methodological insights and limit reproducibility of their research \citep{rogers-closed}. Therefore, the shift towards open source is not only critical for advancing open research and innovation but is also a strong advocacy point for the continued support and investment in open initiatives within the AI community.

\paragraph{\positiveImpact{Open Source Could Be More Affordable}}
\label{sec:near-mid-term-impacts-affordable}

The adoption of Generative AI models is significantly influenced by their cost, impacting both organizations and individuals' decisions to utilize these technologies. A significant factor in the fuelling of growth and wider adoptions of these models in the mid-term is likely to be the productivity gains end-user organizations will see from using Gen AI models (e.g., Buzzfeed layoffs and McKinsey report). These benefits vary with the size of the organization, and will be dictated by cost-benefit analysis. From an individual perspective, Gen AI models can enhance individual productivity by automating repetitive and time-consuming tasks, and augmenting workers when completing more complex and high-value tasks. This can help narrow the productivity gap between workers, improving minimum performance standards \cite{dell2023navigating}.

In principle, open-source models can be more cost-effective than closed source ones, as (1) the model weights are made available for free (if under permissive licenses) and (2) end-users do not have to directly support the high cost of developing these models. Today, a significant barrier to realizing this potential in practice is the complexity involved in deploying and accessing open-source generative AI models. From an organizational perspective, substantial operational costs are still involved, such as the staff required to run the models, the time of leadership to organize and oversee their use, and the compute costs for inference. Some enterprises might also apply additional protections for security and data to ensure compliance when using open-source models, adding further costs. 

According to recent cost comparisons, large organizations with specific requirements benefit from open-source LLMs far more than smaller entities \citep{palazzolo2023meta}. From a user perspective, access to open-source models is currently mostly limited to technically-able individuals. In contrast, closed source models typically have easy to use web interfaces (see \S \ref{sec:near-mid-term-impacts-usability}). Despite these, we are seeing a rise in third-party vendors providing Software Development Kits (SDKs) and API access to a wide range of open-source models (e.g., \href{https://replicate.com/}{Replicate}, \href{https://www.together.ai/}{Together}), which could substantially change the cost dynamics for both individuals and organizations. These third-party vendors can potentially offer lower costs by focusing on building efficient inference infrastructure rather than model development. We anticipate that the cost landscape will evolve in the mid-term, making open-source models more affordable than their closed source counterparts.

\paragraph{\positiveImpact{Open Models are More Flexible and Customizable}}

While Generative AI models are able to perform a variety of tasks out of the box through different prompting strategies \citep{brown2020language}, their capabilities are often limited by the distributions of (1) the pre-training data they were exposed to, (2) the supervised fine-tuning data which was potentially used, and (3) the alignment data and process used  \citep{touvron2023llama2}. As a result, their performance on, for example, low-resource language or very specific tasks falls well below the expectations of developers and users \citep{nicholas2023lost,petrov2023language}. This can often be improved by further fine-tuning \citep{hu2021lora} or alternative techniques such as ensembling multiple model outputs \citep{jiang2023llm}. 

Having access to open-source models, datasets, and assets significantly aids developers in creating models that are high-performing and specifically tailored to their use-case. It also particularly helps cater to less well-resourced languages, domains, and downstream tasks \cite{bommasani2}, as well as enabling personalized models that cater to distinct groups and individuals \cite{kirk2023personalisation}.
These features have created widespread positive sentiment towards open source, which can be seen in venture capital firm's significant investment in open-sourcing efforts \citep{bornstein2023supporting, horowitz}, and the growing adoption of open-source models by companies \citep{marshall2024how}.

\paragraph{\positiveImpact{Open Source Empowers Developers and Fosters Innovation}}
\label{sec:near-mid-term-empower-developers}

Generative AI models with restricted access and control pose significant challenges for innovation. The reliance on third-party providers limits developers' autonomy, hindering maintainability and adaptability as components can be unexpectedly altered or withdrawn. Furthermore, insufficient control over data and data pipelines stifles creativity, making it difficult to innovate, adjust model performance, or fully comprehend workflows. These barriers – often found in closed source models – obstruct the development of tailored, advanced systems, essential for technological progress.

In contrast, open models offer critical advantages to catalyze innovation. Such models empower developers with the capability to tailor the generative AI according to their specific requirements – e.g., by fully controlling the system prompts in LLMs or through fine-tuning –, ensuring a deep understanding and transparency of its mechanisms. This autonomy extends to the management of the data pipeline, significantly bolstering data privacy and the ability to conduct thorough internal audits, an aspect that becomes especially crucial in data-sensitive contexts like schools, government institutions, hospitals, and banks \citep{culotta2023use} (CITE). Note that this is often only fully possible in the event of models that are released under permissive commercial use licenses (see \S \ref{sec:preliminaries}), which is increasingly common in more recent open releases. 

This autonomy and flexibility is particularly pivotal in the burgeoning domain of generative AI-powered agents \citep{chan2024visibility}, where expected outputs involve performing digital or physical actions. Early examples include Adept's ACT-1 model \citep{adepts_post} and Amazon Bedrock \citep{amazon_press_release}. In this context, product developers are likely to value having more control over models, being able to deploy them on-device, and integrate them in larger, more complex systems. Open-source models, with their inherent characteristics of autonomy, flexibility, privacy, and transparency, better equip developers with the capabilities necessary for fostering innovation.

\subsubsection{\SSText{Safety and Security}}
\label{sec:near-mid-term-impacts-safety}

As with any technology, in the near to mid-term Generative AI will bring out new safety and security issues. The primary risks from current and near-term generative AI capabilities comprise two distinct pathways. The first is \textit{malevolent use by bad actors}: individuals or organizations might exploit AI to create damaging content or enable harmful interactions, such as personalized scams, targeted harassment, sexually explicit and suggestive content, and disinformation on a large scale \citep{vidgen2023simplesafetytests,ferrara2023genai}. The second is \textit{misguidance of vulnerable groups}: inaccurate or harmful advice from AI could lead vulnerable individuals, including those with mental health issues, to engage in self-harm \citep{mei2022mitigating,mei2023assert,rottger2023xstest}, radicalize towards supporting extremist groups, or believe in factually inaccurate claims about elections, health, and the environment \citep{zhou2023synthetic}. In the long-term, AI might develop capabilities that present novel existential threats, creating ``catastrophic'' consequences for society such as chemical warfare and environmental disaster \citep{hendrycks2023overview, shevlane2023model, matteucci2023ai}. However, these risks are not a substantial concern for existing models given their limited capabilities. Thus, in the near to mid-term, AI safety primarily means preventing models from generating toxic content, giving dangerous advice, and following malicious instructions. In this work, we are focusing only on the additional benefits and marginal risks to safety from open-source models as compared to closed ones.

\paragraph{\positiveImpact{Open Source Enables Technological Innovation for Safety}}
\label{sec:near-mid-term-impacts-safety-innovation}

One of the foundational principles recognized within the safety community is the inherent impossibility of achieving absolute safety; all systems are susceptible to vulnerabilities, and adversaries will invariably seek to exploit them. For instance, it has been demonstrated that both open-source and closed LLMs are vulnerable to jailbreak attacks that trigger unsafe behaviors \cite{zou2023universal}. Similarly, diffusion-based text-to-image models have exhibited biases, as documented by \citet{nicoletti2023humans}. Consequently, it is essential to continuously foster technical innovation for enhancing safety through diligent research and development in Generative AI systems.

Open source has significantly advanced safety research in the entire model development pipeline. Large open datasets for pre-training, like the Pile \citep{gao2020pile} (released for GPT-Neo, studied in the taxonomy \S \ref{sec:taxonomy}), Laion \citep{schuhmann2022laion}, and RedPajama \citep{together2023redpajama}, can be analyzed for whether they contain toxic content \cite{prabhu2020large}. 
Similarly, open research has shown model fine-tuning to be highly efficient in both improving model safety and removing model safeguards \citep[e.g.][]{bianchi2023safety,qi2023fine}. Unlike closed model APIs, open model analyses permit in-depth exploration of internal mechanisms and behaviors \citep[e.g.][]{jain2023mechanistically, casper2024blackbox}. This transparency enables reproducible and comprehensive evaluations, strengthening our understanding of generative AI safety for models with near and mid-term capabilities. 
Furthermore, open source has also driven innovation in developing safeguards and controls for models, such as Meta's LlamaGuard \cite{inan2023llama} and HuggingFace's \href{https://huggingface.co/spaces/AI-Secure/llm-trustworthy-leaderboard}{Safety Evaluation Leaderboard}.

\paragraph{\negativeImpact{Open Models Can Also be Made to Generate Unsafe Content}}

Generative AI models, both open and closed, harbor the potential to generate unsafe content. This vulnerability spans across the spectrum, with manipulation techniques such as jailbreaks and fine-tuning exposing both types to risks of harmful output. Recent studies have underscored this, revealing that closed models can be just as susceptible to being repurposed for unsafe uses as their open-source counterparts \citep{zou2023universal,qi2023fine}. The universality of this issue necessitates a comprehensive approach to AI safety, emphasizing the importance of both monitoring and innovation in mitigating these risks.

The inherent flexibility of open-source models has been exploited for creating models like \href{https://huggingface.co/ykilcher/gpt-4chan}{GPT4Chan} and various \href{https://huggingface.co/models?search=uncensored}{``uncensored models''} on the HuggingFace hub, designed to bypass ethical guidelines. Yet, this openness also propels advancements in safety technologies, as the open-source ecosystem fosters a collaborative environment for developing safeguards against misuse \citep{dai2023exploring} (see \S \ref{sec:near-mid-term-impacts-safety-innovation}). Overall, the universal nature of these risks justify a more holistic, system-based approach to monitoring unsafe content, rather than focusing on the output of individual models – we discuss this in \S \ref{sec:recommendations}.

\paragraph{\negativeImpact{Open Models Cannot be Rolled Back or Forced to Update}}
\label{sec:near-mid-term-roll-back}

Once an open model is released into the public domain, it essentially becomes immutable in terms of control, with anyone able to download and utilize it indefinitely. It should be noted that this property of open-source models raises potential concerns in several domains (e.g., in the memorization of copyrighted material), but is perhaps chiefly important in safety. While it is theoretically possible to govern the use of such models among compliant users—such as researchers or corporations adhering to ethical norms—through updates to licensing agreements, the reality is more complex. Not all well-intentioned users stay informed about these changes, and those with malicious intent invariably ignore such regulatory attempts. This leads to a significant safety concern, as any vulnerabilities or issues discovered after the model's release cannot be retrospectively fixed or mitigated through model changes. 

In contrast, developers of closed models have the capability to swiftly react to safety concerns by restricting access to compromised models, thanks to their models being distributed and controlled via APIs. To mitigate this risk associated with open models, developers, along with hosting communities like HuggingFace, are urged to adopt rigorous release and access management policies. Following responsible guidelines and the recommendations outlined in previous works (\citealt{solaiman2023gradient, solaiman2023evaluating, anthropic2023}) is crucial for minimizing safety risks and ensuring that the benefits of open-source AI can be realized without compromising safety and security. Moreover, the continuous advancement in generative AI technology, especially in the realm of closed source models which often outpace their open-source counterparts (see \S \ref{sec:taxonomy}, and \citeauthor{uk_gov_safety_2025}, \citeyear{uk_gov_safety_2025}), plays a crucial role in diminishing the attractiveness of older, potentially vulnerable models for malicious use. As these more advanced and capable models become available, they not only offer enhanced functionalities but also operate within ecosystems that are better monitored and governed by the latest safety protocols. This progress suggests that, over time, attackers might find less utility in exploiting older models, given the superior alternatives that are both more effective and subject to stringent safety measures.

\subsubsection{\EAUText{Equity, Access and Usability}}

Generative AI is widely acknowledged for its potential to significantly enhance business efficiency, individual productivity, economic growth, and innovation across various fields, including scientific research. Despite these advantages, critical concerns arise around issues of equity, access, and usability. Predominantly developed by large, Western, for-profit entities, these models spark debates on equity, particularly on who benefits most from this technology. There's a risk that these advancements could exacerbate existing economic and social disparities on a global scale. This challenge is intricately linked to the ease of creating, utilizing, and accessing models tailored to specific needs and contexts. In this section, we explore how open-source models offer considerable advantages in addressing these concerns, providing a more equitable and accessible framework compared to their closed-source counterparts, thereby promising a more inclusive future for generative AI technology.

\paragraph{\positiveImpact{Open Source Increases Model Usability and Accessibility}}
\label{sec:near-mid-term-impacts-usability}

The case of ChatGPT's widespread adoption underscores the significance of user-friendly interfaces in encouraging both organizations and individuals to engage with AI technology. A streamlined, intuitive interface lowers the barrier to entry, allowing users with varying degrees of technical expertise to explore and leverage AI capabilities \citep{scharpnick2023building}. Simultaneously, having easy access to a diverse set of readily available models has immense benefits, allowing users to find the most suitable tools for their specific needs.

Currently, there are few open-source models that have widely-used interfaces matching the one provided by ChatGPT or Gemini. However, emergence of third-party vendors such as \href{https://replicate.com/}{Replicate}, \href{https://www.together.ai/}{Together}, and \href{https://huggingface.co/}{HuggingFace} marks a pivotal shift towards making open-source models more accessible. These platforms provide SDKs, APIs, and downloadable files, simplifying the integration of AI models into various applications. By offering these tools, third-party vendors play a crucial role in bridging the gap between advanced AI research and practical, real-world applications. These third-party vendors are likely to reduce the cost of inference on these models (see \S \ref{sec:near-mid-term-impacts-affordable}), and to make a wide variety of open-source models as easily accessible as ChatGPT is currently to many users.

\paragraph{\positiveImpact{Open Models Could Help Tackle Global Economic Inequalities}}

Knowledge workers in low-income nations, including workers in sectors like call centers and software development, face serious risk of job losses as AI models automate and semi-automate their work. 
Further, if AI models fail to adapt to local contexts, do not work as well for marginalized people and communities, or remain financially inaccessible, the expected economic benefits and new job opportunities may not arise, worsening economic inequalities \citep{imf_blog}.

This is a concern as closed source models are often 1) unaffordable for organizations in low-income countries, and 2) badly-suited to their needs. Local needs are often not met because they lack adequate language support, culturally relevant content, and effective safety measures. This results in higher costs and lower performance, compounding the global inequalities that could be caused by generative AI \citep{petrov2023language, ahia2023all}. In contrast, open models could significantly change this dynamic. With requisite skill building and support for different communities, open models would enable communities to tailor models to their specific contexts and needs, promoting local innovation, safety, security, and reduced bias. This shift could help bridge the growing global inequality gap, paving the way for a more equitable and inclusive future in generative AI.

\paragraph{\positiveImpact{Open Models Can Serve the Needs and Preferences of Diverse Communities}}

To effectively address global needs, it is crucial that models transcend merely reflecting the values of those who are liberal, culturally Western, and English-speaking \citep{aroyo2023dices, lahoti2023improving}. To date, the largest models are primarily developed in the West by for-profit companies, and predominantly trained on Internet data, which is often biased to such demographics \citep{joshi2020state}. This skewness has tangible implications, influencing whom the models serve best and the perpetuation or amplification of detrimental stereotypes and biases within this technology. There is a pressing need for the diversification and enhancement of pre-training datasets to more comprehensively represent a variety of voices and communities. This includes the pursuit of language data collection from underrepresented communities worldwide in a manner that is inclusive of various intersectional identities (e.g., gender), participatory, and non-extractive. Initiatives to increase investment in the development and support of data cooperatives, alongside other frameworks for community engagement and ownership in the data collection and management processes, are particularly promising.

Open-source models offer potential advantages in serving diverse communities. While ongoing efforts are essential to enhance diversity in foundation models' training data, open models, in the shorter term, enable community actors and groups to customize models with datasets that reflect varied contexts, languages, and communities. Open source is a powerful way of achieving this as it helps under-resourced actors build on top of each other’s contributions. For instance, platforms like HuggingFace host a vast array of models, with many designed for specific cultural, geographic, or linguistic needs, e.g., Latxa \citep{bandarkar2023belebele} and LeoLM \citep{hessianai-7b-chat}, covering diverse domains \citep[e.g.][]{li2023chatdoctor}. Closed source models, on the other hand, require the owners of those models to allow for fine-tuning or adjusting the particular models. More research is needed to understand the extent to which fine-tuning open-source models can mitigate pervasive issues of bias in foundation models and better support diverse communities.

\paragraph{\positiveImpact{Open Source Help Democratize AI Development}}

AI systems require specialized and costly resources to develop and deploy them. Therefore, there are limitations in regards to who has the opportunity to develop and deploy them. Many discussions of ‘democratizing’ AI explore how people from different contexts and communities – including those who may lack resources or technical expertise – can be empowered to develop and deploy AI systems. Greater access to build and deploy AI tools can allow for greater innovation, including from more marginalized communities globally. However, ‘democratizing AI’ must also consider the larger picture of how power is held in AI systems and by whom. Large Generative AI models today – both closed and open-source – are developed predominantly by for-profit companies. These models, including their approaches to ‘responsibility’ (or lack thereof), are guided by those companies and internal structures. Thereby, these tools are inherently undemocratic in that tech players own and control their development. 

Open-source models offer some benefits regarding the first interpretation of ‘democratizing AI’. With open source, any developer can hypothetically leverage the investments of larger companies, governments, and research labs, thereby offering to ‘democratize’ AI development and deployment by enabling people to examine, reuse and build on top of these powerful systems. This is vital given the high costs and complexity of developing AI from scratch, from pre-training models, which can cost tens or hundreds of millions of dollars \citep{knight2023openai}, to creating costly human-labeled datasets. This could create societal benefits by enabling broader access and use of AI, which can further innovation and economic opportunities. To realize these benefits of ‘democratizing’ AI development requires commitment to and investments in expanding digital AI literacy, particularly for marginalized communities, as well as skills training for responsible use of open-source models. Further, while open-source models tend to be more transparent than closed models (see \S \ref{sec:taxonomy}, and \citeauthor{bommasani2023foundation}), there is still have much progress to make towards fully ``open'' model pipelines which enable these benefits to be reaped.

It is also important to acknowledge that while open source can enable more access to AI through being able to leverage existing systems, the models themselves still contain key decisions, datasets and approaches that influence what is built on top of them \citep{widder2023open}. In that sense, even open-source models are currently undemocratic. There is significant work required through community engagement, and multi-stakeholder participatory design to integrate broader voices in the design of these systems, ensuring commitments to transparency and explainability.

\subsubsection{\BSAText{Broader Social Aspects}}
\label{sec:near-mid-term-impacts-broader-societal}

The development and deployment of Generative AI models present multifaceted broader societal challenges. Issues encompass public trustworthiness in these systems, copyright complexities surrounding training data, potentially resulting in the models' replication and echoing of such content during inference, as well as environmental considerations which emerge due to the substantial GPU demands. Notably, open-source Generative AI models offer distinct advantages in addressing these concerns compared to their closed-source equivalents.

\paragraph{\positiveImpact{Open Source Improves Public Trustworthiness Through Transparency}}

Nearly three out of five people (61\%) are either ambivalent about or unwilling to trust AI, with \citet{gillespie2023trust} reporting that cybersecurity risks, harmful use, and job loss are the ``potential risks'' that people are most concerned about.
Closed source models pose challenges for evaluating,
benchmarking, and testing them which impede accessibility, replicability, reliability, and
trustworthiness \citep{lamalfa2023language}.
Reports highlighting the discriminatory and biased content produced by currently available models further erodes trust in these systems \citep{nicoletti2023humans,luccioni2023stable}, and the situation is likely to intensify with the advent of embodied Generative AI systems in the mid-term horizon \citep{zhang2023generate}. Transparency is a powerful way of improving trust, and addressing this critical problem. Transparency entails providing detailed and explicit documentation, including provenance artifacts such as model cards, datasheets, and risk cards \cite{gebru2021datasheets, derczynski2023assessing, longpre2023data}. They can be used to assess and review datasets and models, and are widely-used in the open source community. 

Open source is the best way of creating transparency. It enables widespread community oversight as models and datasets can be interrogated, scrutinized, and evaluated by anyone, without needing to seek approval from a central decision-maker. This empowers developers, researchers and other actors to engage with AI and contribute to discussions, encouraging a culture of contribution and accountability \citep{sanchez2021civil}. Simultaneously, the highly technical nature of AI research creates substantial barriers to ordinary citizens. As such, more transparency may not alone drive greater trust -- research outputs also need to be \textit{accessible} and \textit{understandable} by non-experts \citep{Mittelstadt_2019}.

\paragraph{\positiveImpact{Open Models Can Help Reduce Copyright Disputes}}

Generative AI models are at the heart of a growing legal conundrum due to their training on copyrighted material without explicit consent \citep{llmlitigation,metz2024openai}. This issue is exacerbated by the tendency of these models to precisely replicate text from their training data, a phenomenon known as "memorization" \citep{karamolegkou-etal-2023-copyright,carlini2022quantifying}. This practice has led to significant copyright disputes, as evidenced by high-profile lawsuits such as Sarah Silverman vs. Meta and New York Times vs. OpenAI and Microsoft \citep{llmlitigation,metz2024openai}, which revolve around the unauthorized use of copyrighted content in training datasets and the direct reproduction of such texts upon prompting. The opacity surrounding the composition of training data and the challenges in curtailing exact text generation from these datasets underscore a crucial barrier to addressing copyright concerns within the generative AI space.

Open-source models, however, offer a promising avenue for mitigating these copyright issues in ways closed models cannot. By disclosing or detailing their training data (see \S \ref{sec:taxonomy} for the pipeline components), open models can provide much-needed transparency, offering insights into data attribution and potentially clarifying the bounds of "fair use" regarding copyrighted materials. Furthermore, the open nature of these models invites community involvement in refining techniques to reduce memorization, such as through advanced decoding strategies or post-processing methods \cite{karamolegkou-etal-2023-copyright}. Initiatives like the Aya Initiative also demonstrate how open-source approaches can lead to the proactive curation of non-proprietary datasets \cite{hartmann2023sok}, potentially decreasing reliance on copyrighted material and, by extension, the risk of future copyright disputes. This collaborative and transparent model stands in contrast to closed systems, presenting a unique solution to the complex issue of copyright in generative AI.

\paragraph{\positiveImpact{Open Models Can Drive Sustainability in Gen AI Development}}

The environmental footprint of AI model training is a pressing issue, characterized by the substantial energy consumption required for computational resources during both training and inference phases. These processes are significant contributors to CO$_2$ emissions, emphasizing the sustainability challenge within the AI development lifecycle \citep{verdecchia2023systematic, kumar2023how}. The complexity of accurately measuring these emissions arises from variables such as hardware efficiency, operational practices, geographical factors, and the timing of computational tasks \citep{saenko2023computer}. The collective energy demand of the AI industry, highlighted by research like \citep{strubell2019energy} and \citep{wu2022sustainable}, underscores the critical need for strategies to mitigate environmental impacts.

Open-source models offer a promising avenue for reducing energy consumption in generative AI development. By enabling the sharing of pre-trained model weights, reward models, and other artifacts which are resource-intensive to produce, the open-source model approach can significantly lower the requirement for redundant computational efforts across the sector. Furthermore, the transparency inherent in open-source models allows for detailed profiling and optimization of code to identify and rectify energy inefficiencies. This community-driven process facilitates the development of more energy-efficient training methodologies and the exploration of smaller, more sustainable model architectures, as suggested by \citep{schwartz2019green}. Through these collaborative initiatives, open-source AI not only advances technological innovation but also contributes to the pursuit of sustainability in the field.


\subsection{Clarifying ``Real'' Risks and Effective Mitigation Strategies}
\label{sec:near-mid-impacts-real-risks}

This section clarifies important distinctions between ``perceived'' vs ``real'' risks of Generative AI, especially as they relate to open source. These two risks are often muddled in discussions on AI safety in general and with regard to open-source AI in particular. This is similar to how the risk-benefit trade offs of open-source AI are more realistically evaluated within the context of and differential with an alternative closed-source approach rather than as standalone risks.

Two approaches are used here to disentangle ``real'' from ``perceived'' risk. First, several common claims about open-source Gen AI are explored, qualified, and critiqued in detail (\S \ref{sec:near-mid-impacts-claims}). Next, we enumerate the most cited risks associated with AI and closely clarify these risks in terms of the validity of common analogies, existing proven safeguards, and required attacker resource/skills (\S \ref{sec:near-mid-term-impacts-mitigating}). The aim is not to minimize the risks, but rather more accurately qualify and characterize them. A more complete understanding of AI risks not only leads to more accurate risks-benefits trade offs but also helps to develop more effective/focused counter strategies. Like other transformative technologies, our goal is to maximize benefits by using well-established and proven risk management and defense in depth strategies to limit open-source risks.

\subsubsection{Critiquing Common Claims on Open-Source Generative AI}
\label{sec:near-mid-impacts-claims}

\textbf{CLAIM \#1: Closed Models Have Inherently Stronger Safeguards than Open-Source Models.}
Several studies demonstrate that closed models typically demonstrate fewer safety and security risks, compared to open source \cite{rottger2023xstest, chen2024chatgpts, sun2024trustllm}. 
However, closed models still demonstrate weaknesses and, crucially, are also vulnerable to jailbreaking techniques \citep{zou2023universal,chao2023jailbreaking}.
Closed model safeguards are easily bypassed through simple manipulations like fine-tuning via accessible services \citep{qi2023fine}, prompting the model to repeat a word \citep{nasr2023scalable}, applying a cypher \cite{yuan2023gpt4}, or instructing the model in another language \citep{deng2023multilingual,yong2023low}.
Completely preventing models from exhibiting undesirable behaviors might not even be possible \citep{wolf2023fundamental,petrov2024prompting}.
Therefore, it is not clear that closed models are definitively ``safer'' than open-source models. We also anticipate that gaps will narrow over time as jailbreaking attacks become stronger and open safeguarding methods continue to improve. 

\textbf{CLAIM \#2: Access to Closed Models Can Always be Restricted.}
Closed models are often considered more secure because access can be restricted or removed should issues be identified. However, closed models can be compromised via hacking, leaks \citep{llamaLeak}, reverse engineering \citep{appleHack} or duplication \citep{Oliynyk2023stealing}. 
This perspective also assumes that models are only offered through an API. But some closed models are delivered on premise/device, particularly for sensitive deployments (e.g., government applications). In such cases, access may not be retractable.
Finally, closed models can be leaked, e.g., Mistral's 70B parameter was leaked by one of their early customers \citep{franzen2024}. 
Given these factors, developers of closed models do not always have the ability to unilaterally revoke access.

\textbf{CLAIM \#3: Closed Source Developers Can be Regulated to be Safer.}
Regulatory pressure is primarily aimed at large companies building closed source models. For instance, the  \href{https://www.whitehouse.gov/briefing-room/statements-releases/2023/10/30/fact-sheet-president-biden-issues-executive-order-on-safe-secure-and-trustworthy-artificial-intelligence/}{White House Executive Order} required 15 ``leading companies'' to ``drive safe, secure, and trustworthy development of AI.''
Regulatory pressure is a lever for society to create incentives for safe model development. However, regulation is not a panacea and several closed source models have been, and could be, released that are uncensored, poorly safeguarded \citep{elevenScam} or deliberately misaligned \citep{wormGPT,Grok1,Grok2}. 
It is also not clear that regulating closed source models is an effective way of stopping malicious actors \citep{yakuza,NigerianScammers}, who do not comply with regulation and are capable of creating and distributing their own closed source models via illicit sales channels \citep{AiCybercriminal}. 
Instead, regulation might create higher costs for legitimate users who are restricted in what models they can access, develop and share \citep{wu2023llmdet}.

\textbf{CLAIM \#4: All Safety and Security Problems Must be Addressed By the Model Provider.}
It is becoming increasingly clear that, because of the numerous potential applications of generative models, all safety risks cannot be simply identified (and addressed) by the model provider. 
First, most model-related risks depend on access to real-world resources. 
However, legal, financial, and physical constraints significantly hinder activities like acquiring chemicals, equipment and tacit know-how, thus limiting open source's potential for misuse for developing chemical, bio and conventional weapons. 
It is crucial to recognize that effective mitigation of these risks often lies beyond the digital realm. 
Internet Service Providers (ISPs), cloud service providers, social media platforms, along with law enforcement and intelligence agencies, have developed robust measures to mitigate these threats.
This highlights a critical aspect: the potential harm of Gen AI is not determined by access to information but by the capacity to act upon it, which is the same for open and close source models.
Given these factors, safety and security issues cannot be seen as solely the responsibility of the model provider. We expand upon this point in \S \ref{sec:near-mid-term-impacts-mitigating}.

\subsubsection{Mitigating ``Real'' vs ``Perceived'' Generative AI Risks}
\label{sec:near-mid-term-impacts-mitigating}

A systematic error in discussions around the risk-benefit trade-offs of open-source AI is a narrow focus on theoretical risks that is disconnected from real-world implementations and existing safeguards. These theoretical ``perceived'' risks are often presented in terms of familiar but misleading analogies to other dual-use technologies like nuclear energy and synthetic biology. This section enumerates the most common AI threats cited in the literature and media. It pairs them with relevant critiques and qualifiers to more accurately contextualize the ``real'' risk associated for each threat. 

To begin, it is constructive to partition risks into those that are primarily ``informational'' and those that are ``material'' since remediation strategies for each are fundamentally distinct. ``Information AI risks'' are those anchored primarily in the virtual/online world. They manifest as data/information at rest or in motion and are commonly associated with threats like misinformation and intellectual property. ``Material AI'' risks arise in the physical world and include autonomous, nuclear, and/or biological weapons. These material AI risks rely on buying, transporting, assembling, and delivering materials. They often depend on difficult-to-access materials, specialized equipment, restricted environments and unique delivery systems.

This distinction is important because, as a class, material AI risks are generally recognized as much greater threats to humanity. More importantly, the key barriers to effecting material threats are largely unrelated to AI-enhanced information, knowledge, or cognition. Instead, the most difficult aspects of carrying out material AI threats involves acquiring highly controlled and/or costly materials, equipment, delivery systems, etc \citep{Kharaishvili2021}. For example, it is extremely difficult to acquire, access, or otherwise use materials like highly-enriched or even depleted uranium, specialized/automated lab equipment, automation wet labs in the cloud, military grade hardware/components, etc \citep{usnrc2024, canadanuclear2024}. Over the last decades, the two most significant material AI risk areas, nuclear and biological, have evolved stringent multinational safeguards to prevent the acquisition, misuse, and proliferation of key components. \citep{iaea2024,un1540}. Unfortunately, most popular discussions of material AI risks ignore these critical real-world control points. More effective mitigation should focus on  integrating emerging AI models with existing ecosystems and relevant safeguards. \citep{fedamersci2023}.

While the distinction between ``information'' vs ``material'' AI risks is useful, even information AI risks depend upon physical manifestations if only in the form of data and network traffic patterns. The real-world footprints of information risks again afford multilevel, proven defense through in-depth mitigation strategies \citep{NIST2024, ISO27001}. Greater informational threats such as swinging elections via misinformation at scale or instigating cyberwarfare involve more real-world manifestations that are visible in the use of numerous social media accounts, high bandwidth usage, anomalous network patterns, and compute/network infrastructure utilization. Decades of developing sophisticated computer/network security technologies, counter measures, and risk management frameworks can be leveraged as mitigation strategies for the worst AI information risks \citep{cis2021}.

In Table \ref{tab:mitigating_ai_risk} we list several of the risks commonly associated with Generative AI, separating them into Information and Material categories.
The ``Analogy to AI Risk'' column describes how well the ``Threat'' works as an analogy to understand risks arising from AI. Is the risk significantly increased with the availability of open-source generative AI models? For example, material threats as a group are poor analogies since AI does little to overcome the main obstacles for carrying out such threats in the real-world: information is not the bottleneck and such information can often be acquired elsewhere \citep{openaibiorisk2024}. 

\begin{table}[t]
\begin{adjustwidth}{-2cm}{-2cm}
{
\scriptsize
\centering
\begin{tabular}{>{\hspace{0pt}}m{0.125\linewidth}>{\hspace{0pt}}m{0.254\linewidth}>{\hspace{0pt}}m{0.277\linewidth}>{\hspace{0pt}}m{0.32\linewidth}}
\toprule
\textbf{Threat}    & \textbf{Analogy to AI Risk (at scale)}                                                                 & \textbf{Existing Safeguards (at scale)}                                                                             & \textbf{Required Attacker Skills \& Resources}                                                                                                                                       \\\midrulenospacing
\multicolumn{4}{c}{\color{white}\cellcolor{CadetBlue} \textbf{Informational}}\\\hline
Financial Fraud           & \textbf{Bad}: Proven, mature, multilevel safeguards against large-scale coordinated attacks            & \textbf{Excellent}: SOTA multilevel security, constantly evolving with high investment                              & \textbf{Individual}: Low-Medium\par{}\textbf{At scale}: Very High                                                                                                                    \\\hline
Cybersecurity  \par{}Cyberwar   & \textbf{Bad}: Proven, mature, multilevel safeguards against large-scale coordinated attacks            & \textbf{Excellent}: SOTA multilevel security, constantly evolving with high investment                              & \textbf{Individual}: Medium\par{}\textbf{At scale}: Very High                                                                                                                        \\\hline
Disinformation  Fake News & \textbf{Poor}: Proven, mature, multilevel safeguards against large-scale coordinated distribution      & \textbf{Excellent}: SOTA multilevel security, network monitoring and detection                                      & \textbf{Individual}: Low\par{}\textbf{At scale}: Very High                                                                                                                           \\\hline
Deep Fakes                & \textbf{Poor}: Proven, mature, multilevel safeguards against large-scale coordinated distribution      & \textbf{Excellent}: SOTA multilevel security, network monitoring and detection                                      & \textbf{Individual}: Low\par{}\textbf{At scale}: Very High                                                                                                                           \\\hline
Intellectual \par{}Property     & \textbf{Poor}: Cannot exploit IP at scale for profit, but law still evolving                           & \textbf{Good}: Technical monitoring/detection combined with strong IP rights enforcement~                           & \textbf{Individual}: Medium\par{}\textbf{At scale}: Very High                                                                                                                        \\\midrulenospacing
\multicolumn{4}{c}{\color{white}\cellcolor{CadetBlue} \textbf{Material}}\\\hline
Nuclear Weapon            & \textbf{Bad}: Nearly impossible to acquire equipment, secure materials, assemble and deliver materials & \textbf{Excellent}: SOTA multilevel security that restricts, monitors and controls all points in the supply chain~  & \textbf{Individual (or low-profile target)}: Medium\par{}\textbf{At scale (or high-value targets)}: Very High (material acquisition, lab, delivery systems, overcome SOTA defenses)  \\\hline
Bioweapon                 & \textbf{Bad}: Nearly impossible to acquire equipment, secure materials, assemble and deliver materials & \textbf{Very Good}: SOTA multilevel security that restricts, monitors and controls all points in the supply chain   & \textbf{Individual (or low-profile target)}: Medium\par{}\textbf{At scale (or high-value targets)}: Very High (material acquisition, lab, delivery systems, overcome SOTA defenses)  \\\hline
Chemical Weapon           & \textbf{Bad}: Nearly impossible to acquire equipment, secure materials, assemble and deliver materials & \textbf{Very Good}: SOTA multilevel security that restricts, monitors and controls all points in the supply chain~  & \textbf{Individual (or low-profile target)}: Medium\par{}\textbf{At scale (or high-value targets)}: Very High (material acquisition, lab, delivery systems, overcome SOTA defenses)  \\\hline
Dirty Nuke                & \textbf{Bad}: Nearly impossible to acquire equipment, secure materials, assemble and deliver materials & \textbf{Very Good}: SOTA multilevel security that restricts, monitors and controls all points in the supply chain~  & \textbf{Individual (or low-profile target)}: Medium\par{}\textbf{At scale (or high-value targets)}: Very High (material acquisition, lab, delivery systems, overcome SOTA defenses)  \\\hline
Autonomous Weapons        & \textbf{Bad}: Nearly impossible to acquire equipment, secure materials, assemble and deliver materials & \textbf{Very Good}: SOTA multilevel security that restricts, monitors and controls all points in the supply chain~~ & \textbf{Individual (low-profile person)}: Medium\par{}\textbf{At scale (or high-profile targets)}: Very High (coordinated, overcome SOTA defenses, high value targets)               \\\hline
Explosive                 & \textbf{Bad}: Nearly impossible to acquire equipment, secure materials, assemble and deliver materials & \textbf{Very Good}: SOTA multilevel security that restricts, monitors and controls all points in the supply chain~  & \textbf{Individual (or low-profile target)}: Medium\par{}\textbf{At scale (or high-value targets)}: Very High (material acquisition, lab, delivery systems, overcome SOTA defenses) \\\bottomrule
\end{tabular}
}
\vspace{0.5em}
\caption{\textbf{Mitigation Strategies for Perceived AI Threats}: qualitative analysis of existing safeguards and required attacker skills \& resources to carry out threats, divided into \textit{Informational} and \textit{Material} risks.}
\label{tab:mitigating_ai_risk}

\end{adjustwidth}
\end{table}

The ``Existing Safeguards'' column in Table \ref{tab:mitigating_ai_risk} describes the effectiveness of approaches including regulation, monitoring and licensing alongside the difficulty of accessing potentially dangerous components and materials. Finally, ``Required Attacker Skills and Resources'' estimates the skills, knowledge, and resources a bad actor must have to execute a threat in the case that they are able to overcome all existing safeguards. For example, many real-world attacks require substantial financial resources, access to complex/costly infrastructure, and/or specialized knowledge in AI and domains like chemistry or molecular biology. Within the ``Required Attacker Skills \& Resource'' column, a distinction is drawn between carrying out each threat against a relatively low-value ``Individual'' (e.g. random person or someone not under special protection) vs carrying out the threat ``At Scale'' (e.g. against the thousands/millions required to influence public opinion/elections). 

In fact, advanced open-source AI models add little additional risk in the case of cybersecurity/cyberwarfare for two main reasons. First, all current AI models are fundamentally limited by their training data and do not generalize well outside this training distribution \citep{zhang2024}. New, effective security attacks called zero-day exploits are rare, short-lived, and can be valued at over \$100k USD \citep{krebs2016}. There is tremendous incentive to use AI to discover these, assuming it were even possible, and there is no evidence to date to suggest it is. Excluding impractical brute force exploration of combinatorially explosive search spaces, AI models trained on existing threats are poor at `inventing' novel, out-of-distribution security exploits that can't as easily be detected \citep{khandelwal2024}. Second, using variations of existing known exploits within the training data distribution requires an unfocused general scan for susceptible systems that have out-of-date, unsupported, and/or unpatched software/firmware. This would require a large scale random network search that is arguably less effective and easily detected through existing multilevel security measures as compared to focused and tailored attacks by expert human hackers. 

Biosafety is another concerning example that highlights the difference between ``Real'' and ``Perceived'' risks in the context of AI. The potential for misuse of dual-use biotechnology could be devastating and has given rise to a number of international \citep{iso2020, who2021, un2024} and national bio risk safety standards \citep{nih2019, cdc2020, congress2023} over the past four decades. However, AI is even less useful in the context of bioweapons than in the case of cybersecurity. First, unlike the informational nature of cybersecurity, the material nature of biological threats requires purchasing, shipping, and/or accessing a variety of specialized, costly, and/or complex materials that are monitored and do not scale well (e.g. wet lab work). Secondly, the underlying fields of computational biology and quantum chemistry are very complex and less deterministic, thereby making AI-assisted novel predictions and out-of-distribution discoveries extremely difficult.  Given these factors and the fact that most useful distilled information is already accessible elsewhere (e.g. texts, research papers, websites), AI does little to elevate existing biosecurity risks \citep{RAND2017,de2024biosecurity,openaibiorisk2024}.

These two examples on the frontier of AI risk-benefit tradeoffs may change with unforeseen scientific discoveries or future generative AI progress. Our mitigation recommendation is to support the very active multinational and global risk mitigation frameworks and control points with thoughtful incorporation of AI as yet another dual-use technology and tool. Much of this work to incorporate the inevitable progress and proliferation of AI is already been underway for the greatest risks in cybersecurity and biosafety. 

Of special note are AI models and datasets specifically or largely designed as a tool for bioengineering/biosynthesis. These should be integrated into existing biosecurity frameworks for dual-use biological research tools with additional affordances for AI security researchers. AI and biosafety researchers creating such highly-specialized biological tools should responsibly limit access where possible \citep{bran2023chemcrow}.

Like computers and the Internet, AI is a cognitive extension to our ubiquitous global informational infrastructure that brings tremendous benefits. Public, policy, and even research discussion on AI risks in general, and open-source AI in particular, have suffered from three critical flaws. First, there is a singular focus on the risk side of the risk-benefit equation. Second, these popularized risks are mischaracterized and exaggerated, and they appeal to fear. We advocate for more constructive approaches that build upon decades of proven risk mitigation frameworks in critical areas like biosafety and nuclear safety like those cited above. Open-source generative AI can be an essential and valuable compliment to building more effective tools and hardening security (see \S \ref{sec:near-mid-term-impacts-safety-innovation}).

\section{Long-term Impact of Open-Source Models}
\label{sec:long-term-impacts}

Following the discussion of the near to mid-term impacts of open-source Generative AI, we now turn to the opportunities and risks of open sourcing models in the long-term. We defined the long-term phase in \S \ref{sec:prelim-stages} as a period that is enabled by significant new technological breakthroughs that lead to AI with dramatically higher capabilities. For the context of the discussion in this section, we assume this corresponds to achieving some version of what is commonly referred to as \textit{Artificial General Intelligence} (AGI). 

AGI is a contested and somewhat controversial concept, with several competing definitions put forward in previous work. \citet{morris2023levels} define AGI as ``an AI system that is at least as capable as a human at most tasks''; \citet{bubeck2023sparks} define it as ``systems that demonstrate broad capabilities of intelligence, including reasoning, planning, and the ability to learn from experience, and with these capabilities at or above human-level'' and \citet{chollet2019measure} argues that ``an intelligent agent would achieve high skill across many different tasks (for instance, achieving high scores across many different video games). Implicitly here, the tasks may not necessarily be known in advance: to truly achieve generality, the agent would have to be able to learn to handle new tasks (skill acquisition).''

Summarizing what is in common from the various definitions of AGI listed above, we define AGI as \textbf{\textit{a system with high performance capabilities on a wide-range of cognitive or information tasks}}. This definition emphasizes the potential for versatility and effectiveness, without the necessity for physical form.

Although some argue that current AI systems can \href{https://medium.com/@mikeyoung_97230/i-think-youre-testing-me-claude-3-llm-called-out-creators-while-they-probed-its-limits-399d2b881702}{demonstrate situational awareness} and have shown ``sparks'' of AGI \citep{bubeck2023sparks}, it is possible that we never reach such a level of AGI, in which case this long-term phase does not materialize and we stay permanently in the mid-term stage. Because it is uncertain whether we will ever achieve it, it is important to note that discussions of risks and opportunities of AGI are \textbf{highly speculative}. The use of performance and generality levels as done in \citet{morris2023levels} helps operationalize the definition of AGI, but understanding real world impacts of open sourcing the technology requires a wider discussion of the alignment of these systems.

\paragraph*{Technical Alignment} 

Alignment is a key focus of research in AI and robotics \citep{bobu2023aligning}, with generative AI models attracting sustained attention over the past five years as the capabilities of these models have increased \citep{kenton2021alignment}. In an early paper, \citet{leike2018scalable} describe the ``agent alignment problem'' as ``How do we create an agent that behaves in accordance with what a human wants?'' In the simplest sense, this problem describes the challenge of creating models that faithfully act on human instructions to deliver useful outputs (e.g. a model that writes a recipe that can actually be followed). It also describes the challenge of making models that are not harmful and which do not create dangerous side effects. This is critical for responsibly developing and releasing AI models, especially highly capable and easily accessible (open source) ones.

While alignment is desirable in principle, there is no commonly accepted formal definition for it, raising numerous social and theoretical challenges. Most of the concepts involved, such as ``safety'' or ``harm'', are deeply contested and contentious notions, which are difficult to define and operationalize \citep{rottger2021two,morris2023levels}. Aligning a system means answering what \citet{gabriel2020artificial} describes as ``the question of what—or whose—values AI systems ought to align with.'' \citep{gabriel2020artificial}. Yet terminological confusion around ``alignment'' is not necessarily a fatal problem. \citet{kirk2023personalisation} argue that ``while the definition of ‘alignment’ is often vague and under-specified, it is clearly desirable that powerful AI systems, including LLMs, are not misaligned in the sense that they harm human well-being, whether this is through lacking robustness, persuasion, power-seeking, bias, toxicity, misinformation or dishonesty.'' 

Alignment may present methodological challenges, both in terms of eliciting people’s preferences and inscribing them into models. Or, as \citet{hadfield2019incomplete} put it, ``alignment problems also arise because of the difficulty of representing and implementing human values''. More practically, \citet{chun2024informed} demonstrate that LLMs can exhibit latent moral frameworks and ethical reasoning capabilities that do align to some degree with human values. 

For the purposes of our discussion and to distinguish from some of the previous definitions, we define \textbf{\textit{technical alignment}} as the process and outcome of ensuring that AI systems behave in ways that are aligned with the intentions, values, and safety requirements of the \textbf{creators and providers} of the AI system. 

Throughout the rest of this section, we will use this definition to discuss the impacts of open sourcing AGI. We start by tackling an argument that is widely portrayed in the public discourse: that of existential risk. We then move on to discussing alternative scenarios where these \textit{existential} risks do not materialize, discussing benefits and non-existential risks of open sourcing AGI in that setting.

\subsection{Existential Risk and the Open Sourcing of AGI}
\label{sec:long-term-x-risk}

In the context of AI, existential risk (or ‘x-risk’) describes the idea that AGI could lead to human extinction or an irreversible global catastrophe \citep{turchin2020classification,shevlane2023model}. A range of causes have been put forward in prior work, including automated warfare, bioterrorism, rogue AI agents, and unrestricted cyberwarfare that renders technological systems unusable \citep{hendrycks2023overview}. 

Due to the highly speculative nature of AGI, it is impossible to prove or disprove that the probability of human extinction due to direct AGI impacts is non-zero. However even early popularizers of x-risk, like Nick Bostrum, have reversed course on how likely they believe these scenarios to be \citep{bostrom2024deep}. While we believe that existential risk has taken a disproportionate mind-share in the public debate on AI safety \citep{roose2023ai, westerstrand2024talking}, we discuss how open sourcing AI could impact the existential risk that AGI might pose under different development situations.

\subsubsection{Open Source Increases the Likelihood we will Develop Technical Alignment}

In the possibility that AGI is achieved before the development of technical alignment, there is an impending risk of AGI systems becoming uncontrollable.
We hypothesize that under these assumptions the distinction of open versus closed source will start to break down since we do not believe that it would be possible to \textit{contain} AGI systems that are not \textit{controllable}. The x-risk in this scenario is the classical boogeyman story of the rogue AGI that wipes out humanity either out of self-interest or simply because it can.

However, a possible benefit of open-source Gen AI is that it might reduce the probability of this worst-case scenario via three different mechanisms: 
\begin{itemize}
    \item \textbf{Developing and democratizing alignment capacity}. By having open source development of Gen AI, a large number of scientists will be able to work on alignment, as we point out in prior sections. This will radically increase and diversify the talent pool, allowing for faster development of alignment techniques. 
    \item \textbf{Providing early warning}. An open ecosystem of Gen AI will also highlight the risks of unaligned AI systems early on during the development of these systems and thus provide early warnings regarding alignment failures and risks, potentially delaying the creation of unaligned AGI systems. Commercial entities are incentivized to continuously develop the capabilities of the systems, whereas research and academia have focused on evaluating the safety properties, capabilities, limitations, biases and risks of these systems (see \S \ref{sec:near-mid-term-impacts-research} and \ref{sec:near-mid-term-impacts-safety-innovation}). 
    \item \textbf{Democratic monitoring of closed-source model misuse}. Misuse is often considered in the context of open-source models but, as discussed in \S \ref{sec:near-mid-impacts-real-risks}, misuse of these systems can also be done in the closed-source setting, e.g., by entities in jurisdictions with permissive laws or weak rule of law, by governments themselves or by well-resourced criminal organizations. Open-source AGI will enable democratic monitoring of the behavior of closed-source AGI systems, which might help detect and rectify their misalignment.
\end{itemize}

\subsubsection{Conditioned on Technical Alignment, Open Source Helps Maintain the Balance of Power}

Assuming technical alignment, the creators and providers of closed source AI systems can control them in order to accomplish their goals, whether or not these goals are at all aligned with their users or humanity as a whole. This carries significant existential risk, both due to the actions of potential rogue actors within the large entities that build and control the closed source AGI systems, but also due to the incentive structures under which these entities operate (e.g., shareholder primacy and profit maximization).

There are well documented precedents of companies prioritizing profit over the well-being and health of their consumers. For instance, concerns have been raised about the actions taken by cigarette companies, social media companies, and drug manufacturers, all of whom have been alleged to put the well-being of their customers at risk. Further, the long-term safety and well-being of humanity has arguably been jeopardized by accelerating climate change through energy-intensive economic activity. Furthermore, there are alleged efforts of the oil and gas industry to actively obfuscate the risks of climate change.

It is impossible to analyze the risks of alignment failure without considering the balance of power between the public sphere, academia, the civic sector, and the for-profit entities. 

For several of the aforementioned problems, there currently exist a large number of mechanisms to limit the worst failure modes of unchecked profit-seeking by powerful corporations. However, it is inconceivable that this \textit{balance of power} would be maintained if a technically aligned closed-source AGI was introduced by a corporation or single government. Clearly, this type of AGI would provide unprecedented powers to the entities that control it. Thus, it would be very difficult for other actors to maintain effective oversight and governance at global scale, especially if the systems were not publicly announced.

In contrast, with the release of open-source models, the balance of power is more likely to be maintained, since all different parties are able to use AGI systems to advance their own interests and keep each other in check. For example, the regulators would be able to use the systems to keep in check the corporations, and consumers could use these systems to protect their own interests. In other words, by making AGI systems open source different actors can align these to their own interests.

\subsubsection{Open-Source AGI Might Help Us Develop Better Decentralized Coordination Mechanisms}

Enhancing the likelihood of technical alignment and preserving the current balance of power can mitigate existential risks; however, it is evident that our existing systems for maintaining this balance and resolving conflicts are significantly flawed. For example, the inability of coordinating collective action on a global scale is a major hindrance in our ability to address climate change, environmental damage and global inequality. Similarly, even in the 21$^{\text{st}}$ century our inability to efficiently resolve conflicts has resulted in large scale wars. 

Instead of simply maintaining and stabilizing the current mechanisms of balance of power, open-source AGI, under technical alignment, might be able to help develop better mechanisms of coordinating societies and resolving conflicts of interest. While this kind of technology is out of reach given the current state of alignment research, we believe that this is ultimately the only path to AGI that has the potential to guarantee safety of these systems. However, we note that its potential to strengthen democratic structures can only be realized through an open community effort, which is fundamentally incompatible with closed source. 

\subsection{Benefits and Non-existential Risks of AGI and Open Sourcing Systems}
\label{sec:long-term-benefits-non-x-risks}

While a significant portion of the public discourse on the risks of AGI tends to focus on the existential variant discussed in the previous section, there are unique benefits and non-existential risks that the technology provides. 

\subsubsection{Non-Existential Risks of AGI and Open Source as a Mitigation Strategy for Some Risks}

The combination of recent trajectories of socio-technological phenomena guided by established historical patterns enables a productive discussion on AGI risks from broader perspectives incorporating individual, social, cultural and economic concerns. In particular, this long-term perspective provides another opportunity to contrast open vs. closed source approaches to AGI. It also illustrates how such decisions today can influence the shape of future society. 

Some of the most popularly discussed and/or impactful long-term AI risks are listed in Table \ref{tab:non_x_risks_agi}. Each risk is labeled by the dominant character (Culture, Social, Individual, Economic) with a ``Description''. Risks are grouped into either ``Common'' or ``Enhanced by Closed Source AGI'' to contrast (a) common convergence towards universal future risks against (b) elevated long-term risks due to closed-source approaches. This division is somewhat fuzzy. The unrestricted proliferation of increasingly capable AI in a \textit{laissez-faire} environment could lead to long-term risks common to both open and closed source that are distinguished primarily by differing rates of development. 

In context of these long-term risks, the main advantage of open-source AGI is the democratizing effects it enables, starting with broad-based research into critical issues like bias, transparency, explainability, safety, alignment, etc. This, in turn, both informs and enables more diverse experts to collaborate on more effective mitigation strategies for these risks. 

\begin{table}[t]
\begin{adjustwidth}{-1cm}{-1cm}
{
\scriptsize
\centering
\begin{tabular}{>{\hspace{0pt}}m{0.19\linewidth}>{\hspace{0pt}}m{0.1\linewidth}>{\hspace{0pt}}m{0.69\linewidth}}
\toprule
\textbf{Threat}                 & \textbf{Risk Type} & \textbf{Description}\\
\multicolumn{3}{c}{\color{white}\cellcolor{CadetBlue} \textbf{Common}}\\\hline
Cultural \par{}Homogenization         & Culture            & Globalized hypercompetition, economies of scale, and prioritization of the same profitable OECD markets are strong economic incentives for the handful of massive SOTA commercial AGI models to be trained, tuned, aligned, and embedded in similar ways, constrained by increasingly limited high-quality data, and exhibiting similar human-AGI alignments and latent knowledge, reasoning, and ethics.                                          \\\hline
Impoverished Social \par{} Connections & Social             & Ubiquitous, performant, and emotionally intelligent AGI trained to accommodate human needs may be more desirable than trade-offs involved in human-human social interactions.                                                                                                                                                                                                                                                                      \\\hline
Mass Unemployment               & Economic           & The world's largest/wealthiest corporations are narrowly incentivized to maximize shareholder value and will focus on the huge economic opportunity for mass automation via AGI.~                                                                                                                                                                                                                                                                  \\\hline
Inequality                      & Economic           & AGI can cannibalize traditional human knowledge work and physical labor. This aspect of AI technical disruption inherently concentrates wealth from the broader economy.~                                                                                                                                                                                                                                                                          \\\midrulenospacing
\multicolumn{3}{c}{\color{white}\cellcolor{CadetBlue} \textbf{Enhanced by Closed Source AGI}}\\\hline
Cultural Bias                   & Culture            & The democratizing effects and access to open source AGI enables diverse groups to monitor, create, and preserve their interests beyond the mainstream.                                                                                                                                                                                                                                                                                             \\\hline
Social Manipulation             & Social             & Debate, marketing, persuasion, deception, etc. all lie along an ethical spectrum of un/acceptable rhetoric. Open source seeds research, informs oversight, and can guide policy discussion to help develop ethical operational boundaries.                                                                                                                                                                                                         \\\hline
Mental Illness                  & Individual         & Open source AGI can help individuals understand, critique, and develop healthy coping strategies for both social and technological stressors we see. This includes AI assistants, therapy bots and more healthy/pro-social social media that are independently aligned toward individual mental health.~                                                                                                                                           \\\hline
Cognitive Atrophy               & Individual         & AGI increasingly automates both traditional cognitive tasks as well as seemingly creative tasks. Profit-focused tech companies historically prioritize customer capture, dependence, and recurrent revenue via proprietary closed gardens. Open source AI can alleviate this by providing more transparency, customization, and human-in-the-loop engagement that reduces blind-faith dependence on closed-source vendors.                         \\\hline
Diminished Intimacy             & Individual         & For better or worse, humans are more atomized than ever using traditional metrics of close friendships, dating, marriage, sex, etc. By some engagement metrics already, humans prefer to engage with online chatbots like character.ai, pi.ai and xiaoice over other humans. With advances in AI, especially embodied AI, closed source for-profit AI may be able to accelerate this dependency without ethical or informed regulatory oversight.  \\\hline
Eroded Autonomy                 & Individual         & As a system, humans present a large attack surface for AGIs to manipulate. This is why the EU AI Act specifically calls out safeguarding ‘Human Autonomy’. Closed source betrays inherent conflicts including profitability.Open source provides more opportunity for widespread collaboration on prevention given the interdisciplinary nature of the problem and the fundamental importance of preserving human agency and dignity.~             \\\hline
Neo-Feudalism                   & Economic           & AGI will likely have wide moats in terms of capital, infrastructure, human expertise, and large-scale data that strongly favor the largest players and first movers. As the long-term advantages accumulate, power concentrates and market efficiencies/mechanisms like pricing/regulation begin to break down. Open-source is one of the most important counter measures to organically preserve market efficiencies without excess.\\\bottomrule
\end{tabular}
}
\vspace{0.5em}
\caption{\textbf{Non-Existential Risks from AGI}: categorized into \textit{Common} risks from open and closed source models, and risks that are \textit{Enhanced by Closed Source AGI}.}
\label{tab:non_x_risks_agi}

\end{adjustwidth}
\end{table}

To illustrate one common long-term risk, consider mass unemployment. Recent studies suggest college educated, high paid knowledge workers in developed countries face the greatest risk of AI automation given the relative return on investment (ROI) and current AI model strengths \citep{ellingrud2023generative,hatzius2023potentially}. These job categories include analysts, programmers, business writers, graphic designers, among many. While there are debates over rates of progress and suggested ameliorative public policies like universal basic income, these don’t negate the underlying reality. Can economies create as many new and economically essential (e.g. lucrative) jobs to maintain the balance of creative destruction caused by AI technological disruption? It could be argued that open-source can mitigate the risk of massive unemployment by creating a more decentralized, competitive market with greater likelihood to discover and fund new job opportunities. Nonetheless, this is a more speculative argument with less historical evidence. For this reason, mass unemployment AI risk is categorized as a risk common to both open-source and closed-source approaches to AI.

In contrast, social manipulation is categorized as a long-term risk that is comparatively higher with closed source. These range from risks to consumers to risks to democracies given large-scale election interference. There are also significant mental health risks to large populations. Moreover, all of these risks have the potential to become interrelated. Mitigating such social risks requires collaboration across disparate disciplines from identifying human susceptibility to determining ethical and legal consequences and recommending mitigation strategies. Domain experts from varied fields such as psychology, cognitive science, neuroscience, behavioral economics, philosophy, law, marketing/sales, and political theory can all help design solutions.

A closed-source oligopoly among a handful of the world’s largest corporations dramatically increases risks to humanity by subjugating all considerations to shareholder maximization. The hypercompetitive race among leading AI corporations creates market incentives for tactics favoring exploiting social manipulation and addiction. Opaqueness is favored to decrease corporate liability. This misaligns AI along narrow business interests instead of aligning it for broader social good. 

Closed-source AI policies may also lead to more socially deleterious political lobbying, regulatory capture, and revolving doors with high government officials, especially in the US where such practices are common. In contrast, open-source can act as an invaluable secondary market of diverse, self-interested stakeholders that can provide greater transparency, safety, and accountability.

\subsubsection{Benefits of Open-Source AGI}

The potential benefits of AGI are often ignored in favor of the discussion of the risks associated with the technology, yet they are substantial to both humanity and individuals. We believe the community should be at least as concerned about the potential missed opportunities of open-sourcing AGI as it is about the potential existential risks of it. These can include the economic benefits detailed above, as well as mitigation strategies to prevent social manipulation. Benefits also extend to the democratizing effects as diverse groups are able to  create and preserve their interests, as well as the possibilities for increased development in AI solutions to mental health, healthcare medicine and scientific R\&D.  In addition to the many comparative advantages outlined in Table \ref{tab:non_x_risks_agi}, open-source would allow for greater participation in developing AI solutions like clean energy, as well as advances in education, transportation and mobility. Open-source offers additional opportunities for developing guardrails for safe AI, but it also increases potential contributions to other areas of public safety and security. While much focus has been on the negative impact on writers and artists, some have theorized new contributions in the arts, culture and entertainment \citep{latif2023artificial,whitehouse2022impact,gruetzemacher2022transformative}. 

Crucially, many of these benefits will only be fully realized under the umbrella of open-source AGI, following a similar discussion to the one in the near to mid-term section (\S \ref{sec:near-mid-term-impacts}). Under closed source AGI, many of these benefits could turn into the risks under the framework previously discussed. For example, economic growth benefits under closed source AGI could be extremely concentrated and lead to major growth in economic inequality risk, eroded authority and neo-feudalism. Ultimately, open sourcing will allow for several of the benefits to be reaped while mitigating many of the potential risks associated with the transformative technology.

\section{Recommendations for Policy and Best Practices}
\label{sec:recommendations}

Based upon our analysis of the opportunities and risks of open-source generative AI in \S \ref{sec:near-mid-term-impacts} and \ref{sec:long-term-impacts} and the limited marginal increase in risks compared with the proprietary-only baseline (\S \ref{sec:near-mid-impacts-real-risks}), we make the following recommendations with respect to the open-sourcing of generative AI models:

\begin{quote}
In order to foster the innovative and creative development of Gen AI as well as the broad-based testing and vetting of new models to prevent biases and harm, we strongly favor appropriate legislation and regulation of the improper \textbf{use} of generative AI models, and we would be happy to work with legislators and regulators to develop such provisions.
 
At the same time, we believe it is in the best interest of society, especially marginalized communities, to \textbf{not restrict the open-source development of generative AI} by ensuring developers are not liable for the improper or illegal use of the resulting models provided that their models as developed do not encourage such misuse.
 
An exception to this proposed policy should be developers who obtain training data or who produce models reasonably known or suspected to be illegal at the time of the model’s development. In our view, this approach reaches the optimal balance between the opportunities and risks associated with open sourcing Gen AI models.

\end{quote}
 
Given the black-box nature of generative AI models and the need to include a wide and diverse set of researchers and communities, we believe open-source models allow the largest number of participants to be involved in identifying and mitigating potential harms. The transparency and availability of open-source models will continue to provide the most widespread access to large-scale monitoring and auditing post-release. This is especially important for underrepresented communities across the world who may not otherwise be included in the development and evaluation of these models.

We discussed in detail the opportunities and risks of open sourcing Gen AI models and systems in \S \ref{sec:near-mid-term-impacts} and \ref{sec:long-term-impacts}. In particular, open-source models are more flexible and customizable, empower developers and foster innovation, advance research, improve trustworthiness and transparency, and can be more affordable and accessible than closed source ones. They also hold the potential of democratizing the development of AI, and help meet the needs of diverse communities, with crucial implications in terms of balance of power in the long-term horizon (see 
\S \ref{sec:long-term-impacts}).

Some have argued that open-source Gen AI models could introduce direct safety risks and concerns by empowering malicious actors and users. However open-source Gen AI models have been, and we expect they will remain, at the forefront of safety, fairness and equity research and are hence indispensable for the safe development and deployment of Gen AI systems.

The key question is whether the benefits of open sourcing Gen AI models outweigh the marginal increase of the direct safety and security risks beyond keeping them fully closed source. 
We argue that in the \textit{near to mid-term} open sourcing introduces only a marginal increase of the immediate risk compared to the closed source setting (\S \ref{sec:near-mid-impacts-real-risks}), and that in the \textit{long-term} open-source models are key to avoiding existential risk (\S \ref{sec:long-term-x-risk}) and helping to maintain a healthy balance of power as well as reducing other non-existential threats (\S \ref{sec:long-term-benefits-non-x-risks}). 
At this point, it is important to recognize closed source models are not a silver bullet in terms of safeguards (e.g., jailbreaks and leaks have occurred in closed source models). As such, it is key to address safety concerns in the real-world context of their deployment rather than through the model's access level.
In fact, restricting open sourcing can incur profound missed opportunities for economic development and safety research, especially as generative AI affects disempowered communities around the world. 

We believe benefits and the risk of the missed opportunities of open sourcing Gen AI \textbf{significantly outweigh} the marginal direct safety and security risks. Hence, we advocate against introducing restrictions on the development of open-source generative AI models while encouraging regulation of the improper use of such models. 

Importantly, we accept that there may be residual and unpreventable harms to individuals and communities caused by systems using generative AI systems, including such based on open-source models. Such harms may not be preventable or rectifiable with technical means. Therefore, we also strongly advocate for comprehensive societal mechanisms to support such individuals and communities.

Many of the risks of open-source models can be largely mitigated with voluntary commitments and best practices. We want to stress the importance of \textbf{keeping these practices voluntary and not regulating open-source development}, as regulation and legal uncertainty carry the risk of suppressing the fragile open-source AI ecosystem. In particular, we recommend that open-source developers, whenever practical, relevant, and proportional, consider:

\begin{enumerate}
    \item \textbf{Pre-Development Engagement:}
We recommend that developers start engaging critically with potential stakeholders to discuss the broader impacts their models may have well before they even start developing them, and consider whether the models should be developed in the first place \citep{bender2021stochastic} as well as whether the model would eventually be open sourced. We urge model developers to consider the impact of the models and datasets on broader communities, to put as much effort as reasonable in debiasing their training data, increasing coverage of content and languages, especially for less represented groups, as well as maximizing the accessibility and utility of their models and tools for as many individuals and organizations as possible. We also recommend engaging with ethics specialists, such as institutional Ethical Review Boards, especially in cases where the model can affect individuals’ and communities’ wellbeing before, during, or after model development or open-sourcing. Ideally, representatives of the affected communities should be involved at every step of this process.
 
    \item \textbf{Training Transparency:}
As illustrated by Table \ref{tab:classification}, a lack of data transparency is a problem even in relatively open LLMs (e.g. LLaMA-2 or Mistral). Making training and evaluation data publicly available enhances the community’s capacity to scrutinize models’ capabilities, risks, and limitations, thereby unlocking many of the advantages outlined in \S \ref{sec:near-mid-term-impacts}. A good step in this direction are Pythia \citep{biderman2023pythia} and StableDiffusion \citep{Rombach2022StableDiffusion} whose training data is released alongside the models. This has already enabled research into memorization \citep{ippolito2023preventing,biderman2024emergent}, privacy of the training data \citep{li2023mope,duan2024membership}, understanding sources of bias \citep{seshadri2023bias,belem2024are} and studying how to prevent unsafe generations \citep{Schramowski2023SLD,brack2023mitigating,zhang2023generate}. Therefore, we recommend that other open-source model developers also release their training and (safety) evaluation datasets alongside training, inference and evaluation/benchmarking code, intermediate checkpoints and training logs. We believe this will enable further safety research which requires more parts of the pipeline to be open. 
 
    \item \textbf{Safety Evaluations:}
When developing new general-purpose models or building on existing ones in ways relevant to their safety performance, we recommend following the best practices for safety evaluation available at that time. 
Often this can be achieved by adopting recent tried-and-tested industry-level safety benchmarks \citep{vidgen2024introducing}. Where viable, a thorough safety evaluation should encompass both manual and automated testing, ranging from adversarial jailbreak prompts to expert-led red-teaming for common and edge case exploits.
However, developers should recognize that safety practices should evolve alongside Gen AI development, and we should not expect past practices to be sufficient. As such, there should also be some proactive action to create novel safety practices.
%
Presently, even among the model developers who perform safety evaluations, few open-source their safety evaluation pipelines, as seen in Table \ref{tab:classification}. This is an issue as it leads to safety evaluation lagging behind model development. Hence -- following the above transparency recommendation and aiming to enable a safety benchmarking culture -- we urge developers to also open-source their safety evaluation datasets and evaluation code. 
 
    \item \textbf{Documentation:}
Responsible open-sourcing does not only pertain to the model weights but also includes appropriate accompanying artifacts \citep{mixtral_tweet}.
To this end, we recommend model developers provide clear documentation on how the model was developed and tested, what use-cases it targets, how well it can be expected to perform, and what considerations downstream users should take. A discussion on the broader implications of developing and releasing the model -- similar to the Ethical Statements required at the major AI conferences -- would also be beneficial. For models that could potentially be used for illicit downstream purposes, we also recommend that model developers consider explicitly providing licensing restrictions on certain uses of that model.

\end{enumerate}

With these recommendations and best practices, we believe it is possible to mitigate the risks and address the prevalent concerns with open sourcing these models, paving the way for realizing the vast potential benefits open-souce Gen AI has to offer.

\section{Related Work}
\label{sec:related-work}

While there have been a flurry of reports and surveys on the impact of general open source software in areas such as innovation or research within the last few decades \citep{rossi2012adoption,jaisingh2008impact,brown2002reusing,von2007open,ebert2007open}, the discourse surrounding the openness of generative AI models presents unique complexities due to the distinctive characteristics of this technology. To address this, we divide this section into two areas of related work: one examining the broader impact of generative AI technology, and another focused on the specific debate surrounding open sourcing these models.

\subsection{Impacts of Generative AI}

The studies of the influence of generative AI technology in various aspects of society and ethics can be broadly categorized into two camps: those dealing with benefits and risks of the technology as it exists today, and those focusing on the potential impacts of a capability shift. Within the first category, there are a variety of studies on the expected effects of generative AI within specific areas of application, such as science and medicine \citep{ai4science2023impact,fecher2023friend,birhane2023science}, education \citep{alahdab2023potential,cooper2023examining,malik2023exploring,susarla2023janus}, the environment \citep{rillig2023risks}, the labor market and the economy \citep{eloundou2023gpts,rotman2023chatgpt,brynjolfsson2023generative,chui2023economic}, copyright \& fair use \citep{henderson2023foundation,samuelson2023generative}, law enforcement \citep{europol2023chatgpt}, information \citep{simon2023misinformation}, cybersecurity and privacy \citep{gupta2023chatgpt}, amongst others, as well as more general on society \citep{toumi2021functorial,tokayev2023ethical,baldassarre2023social,saetra2023generative,sabherwal2024societal}. 

There is a large body of work on the potential impacts of capability shifts in generative AI \citep{sastry2021beyond,liang2022time,solaiman2019release}. Recent research evaluates the associated benefits and risks from a variety of perspectives. \citet{seger2023open} explore proliferation dangers but also advocate for judicious openness to enhance safety. The link between release policies and issues of oversight and access is a fundamental point in other studies. \citet{kapoor_narayanan_2023} emphasise the critical importance of transparently analysing AI failures to ensure its safety. From a different perspective,~\citet{widder2023open} suggest that current practices do not sufficiently enable public scrutiny and participation. While much research concentrates on access to model capabilities,~\citet{henderson2023foundation} and~\citet{chan2023hazards} raise concerns that increasingly accessible fine-tuning of downloadable models may amplify hazards and undermine existing safety mechanisms.


While the aforementioned works in this section correspond to important previous research in this area, the focus is mostly on the impact of generative AI. In contrast, our work focuses on primarily analyzing the distinctions between open and closed generative AI models in terms of their current/near, mid and long-term impacts.

\subsection{On Open Sourcing Generative AI Models}

While closed-source Gen AI models still exhibit superior performance compared to their open-source counterparts~\citep{bommasani2021opportunities}, this gap is steadily narrowing~\citep{chen2024chatgpts,patel2023google,uk_gov_safety_2025}. This is prompting an important debate regarding the optimal practices for open releases of such systems, to mitigate associated risks.

One main line of discussion is centered on the categorisation of these systems, based on their agreement to open disclosure of training pipeline, weights, and data~\citep{bommasani2023foundation,liesenfeld2023opening,seger2023open,bommasani2,shrestha2023building,widder2023open}. The term "open-source" itself may not be entirely appropriate for AI systems~\citep{Lukianets,maffulli}, which typically encompass more than just code, hence necessitating custom release pipelines~\citep{liu2023llm360}. A considerable number of opinion articles~\citep{laion,hacker2023regulating,tumadottir} highlights the need to differentiate open-source systems from closed-source ones from a regulatory standpoint. This distinction is crucial to ensure that open-source contributors do not face unsustainable compliance costs in light of regulations such as the EU AI Act~\cite{euact}.

Another line of work is focused on the aforementioned AI safety, which acquires further importance in the context of open source AI, due to the limited control obtainable once models are released. Many have raised the need of open models for mitigating the risks of centralisation~\citep{seger2023open,horowitz,shrestha2023building}, while improving developments and facilitating further research on the topic~\citep{kasneci2023chatgpt,arstechnica,spirling2023open}. On the other hand, open models may logically exacerbate the risks of misuse of generative AI~\citep{bommasani2021opportunities,alaga2023coordinated}. Interestingly, it has also been shown that open Gen AI tends to be less trustworthy than closed one~\citep{sun2024trustllm}. This correlates with the different perception of the impact of potential misuse in open source contributors~\citep{widder2023open}. A relevant paper~\citep{seger2023open} attempts to draw conclusions about the risks and benefits of open models, and shapes recommendations for the near future.

In our work, we provide a novel viewpoint on both research directions, in a holistic manner. We propose a new taxonomy for distinguishing open models, and a novel distinction of risks associated with the near, mid and long-term future, finally providing related recommendations. This work is an extension of \citep{eiras2024near}, expanding on the near to mid-term impacts, providing an analysis of the long-term risks and opportunities, and proposing more thorough recommendations to developers.

\section*{Acknowledgments}
The authors would like to thank Meta for their generous support, including travel grants and logistical assistance, which enabled this collaboration, as well as for the organization of the first Open Innovation AI Research Community workshop where this work was initiated. Meta had no editorial input in this paper, and the views expressed herein do not reflect those of the company.

FE is supported by EPSRC Centre for Doctoral Training in
Autonomous Intelligent Machines and Systems [EP/S024050/1] and Five AI Limited.
AP is funded by EPSRC Centre for Doctoral Training in
Autonomous Intelligent Machines and Systems [EP/S024050/1].
FP is funded by KAUST (Grant DFR07910).
JMI is funded by National University Philippines and the UKRI Centre for Doctoral Training in Accountable, Responsible and Transparent AI [EP/S023437/1] of the University of Bath.
PR is supported by a MUR FARE 2020 initiative under grant agreement Prot. R20YSMBZ8S (INDOMITA).
PHST is supported by UKRI grant: Turing AI Fellowship EP/W002981/1, and by the Royal Academy of Engineering under the Research Chair and Senior Research Fellowships scheme.
JF is partially funded by the UKI grant EP/Y028481/1 (originally selected for funding by the ERC). JF is supported by the JPMC Research Award and the Amazon Research Award.

%
%
%
\bibliographystyle{acl_natbib}
\bibliography{refs}
%





\end{document}